\documentclass{article}

\usepackage{microtype}
\usepackage{graphicx}
\usepackage{stfloats}
\usepackage{caption}
\usepackage{subcaption}
\usepackage{booktabs} % for professional tables
\usepackage{amsmath}
\usepackage{amsfonts}

\newcommand{\comment}[1]{}
\usepackage{hyperref}

\usepackage[accepted]{icml2021}

\icmltitlerunning{Understanding the effect of varying amounts of replay per step}

\begin{document}

\twocolumn[
\icmltitle{Understanding the effect of varying amounts of replay per step}
\icmlsetsymbol{equal}{*}
\begin{icmlauthorlist}
\icmlauthor{Animesh Kumar Paul}{equal,uofa}
\icmlauthor{Videh Raj Nema}{equal,uofa}
\end{icmlauthorlist}
\icmlaffiliation{uofa}{Department of Computing Science, University of Alberta, Canada}
\vskip 0.3in
]
\icmlcorrespondingauthor{Animesh Kumar Paul}{animeshk@ualberta.ca}
\icmlcorrespondingauthor{Videh Raj Nema}{nema@ualberta.ca}

\printAffiliationsAndNotice{\icmlEqualContribution} % otherwise use the standard text.

\begin{abstract}

Model-based reinforcement learning uses models to plan, where the predictions and policies of an agent can be improved by using more computation without additional data from the environment, thereby improving sample efficiency. However, learning accurate estimates of the model is hard. Subsequently, the natural question is whether we can get similar benefits as planning with model-free methods. Experience replay is an essential component of many model-free algorithms enabling sample-efficient learning and stability by providing a mechanism to store past experiences for further reuse in the gradient computational process. Prior works have established connections between models and experience replay by planning with the latter. This involves increasing the number of times a mini-batch is sampled and used for updates at each step (amount of replay per step). We attempt to exploit this connection by doing a systematic study on the effect of varying amounts of replay per step in a well-known model-free algorithm: Deep Q-Network (DQN) in the Mountain Car environment. We empirically show that increasing replay improves DQN's sample efficiency, reduces the variation in its performance, and makes it more robust to change in hyperparameters. Altogether, this takes a step toward a better algorithm for deployment.

\comment{
Sample efficiency is critical for real-world applications of deep reinforcement learning. Experience replay helps improve sample efficiency and stability by providing a mechanism to store past experiences for further reuse in the gradient computational process. One of the recent works shows that it is possible to improve the sample efficiency of Deep Q-Networks (DQN) further by increasing replay per step, which is one of the experience replay hyperparameters, and at the same time, it could help to provide competitive performance for DQN compared to a model-based agent that uses the planning step. However, they did not do a systematic study investigating sample efficiency with different values of replay per step and their effect on the algorithm’s sensitivity to other hyperparameters. This motivates us to empirically study the effect of varying the amount of replay per step. We empirically show on the Mountain car environment that the sample efficiency of DQN improves by increasing the number of times a mini-batch is replayed per step. We also found that the DQN agent with higher replay per step is less sensitive than the standard DQN in terms of selecting the hyper-parameters values.}

\end{abstract}

\section{Introduction}
\label{introduction}

Deep Reinforcement Learning (RL) has demonstrated its competence to solve sequential decision-making problems ranging from games~\cite{Mnih2015, Silver2016}, simulated environments~\cite{Mnih2015, Schulman2015}, finance~\cite{rlfinance}, and robotics~\cite{Levine2016}. These advancements have been achieved at the cost of huge computational requirements and a large number of interactions with the environment. A promising approach to having sample efficient learning is to reuse valuable past experiences to limit the collection of new experiences from simulation or the real world. This need for sample efficiency can be handled by Experience Replay~\cite{Lin1992}. 
% Wang2017, Zhang2017

Experience Replay (ER) has been widely adopted in deep Q-learning due to its data efficiency and the stability it induces~\cite{Mnih2015, Schaul2016, Hessel2018}. ER works by storing the agent's experiences (transitions) in a fixed-size buffer and sampling a mini-batch of them at each step to update the neural network (function approximator), thereby mixing the more and less recent experiences for learning. Several modifications of ER have been proposed in terms of non-uniform buffer sampling or experience re-weighting~\cite{Schaul2016, DeBruin2018, Sinha2022}. There has also been some work on replaying experiences with a goal different from what the agent was trying to achieve to learn even from undesired outcomes (Hindsight ER~\cite{Andrychowicz2017}).

\comment{It not only improves sample efficiency but also breaks the temporal correlation between samples which helps to conserve the i.i.d. assumption of stochastic gradient-based algorithms~\cite{Bruin2015}.
The basic working principle of ER is that we store the most recent experiences (transitions) in a fixed-size buffer, and sample a mini-batch of transitions from it at each update step. Hence, ER mixes more and less recent experiences for learning with neural networks. As a result, we can make multiple updates to the network using rare previous experiences, which might become useful later in the learning phase when the agent's competence is increased ~\cite{Schmidhuber1991}. ER stabilizes and improves the network training process for learning value functions~\cite{Mnih2015}. Subsequently, There has been an increase in the application of ER for deep RL.} 

The ER mechanism embeds several hyperparameters like the size of the buffer (capacity), the replay mini-batch size, the number of transitions to store before starting updates, and the amount of replay per step. Some noteworthy studies have been conducted to empirically measure the importance of tuning these hyperparameters across a variety of deep RL algorithms. For instance, it has been shown that the performance of deep RL algorithms degrades due to a large replay capacity~\cite{Zhang2017}. Further, it has also been identified that increasing the replay capacity and reducing the effect of the oldest policy in the buffer (transitions collect by that policy) improves the performance~\cite{Fedus2020}. In this work, we attempt to investigate the variation in algorithm behavior due to a change in one of these hyperparameters: the amount of replay per step. Typically, a mini-batch of transitions is sampled from the ER buffer only once per training step~\cite{Mnih2015}. However, we can put a loop around this by sampling a mini-batch multiple times. This notion is referred to as the amount of replay per step or the \textit{replay frequency}, where a replay frequency of $\tau$ implies sampling a mini-batch and using it to update the network parameters sequentially $\tau$ times per training step.

The motivation for varying $\tau$ comes from the work done by van Hasselt et al. ~\cite{VanHasselt2019}, where they make connections between parametric models and ER. Model-based algorithms use models to \textit{plan}, i.e. using more computation to improve the predictions and policies without consuming additional data in the form of new interactions with the environment. They argue that the experience stored in the ER buffer of a model-free algorithm (like DQN) can be similarly used to plan. They provide empirical evidence in Atari where Rainbow DQN (a variant of DQN~\cite{Hessel2018}) with experience planning (i.e. increased $\tau$) achieves better sample efficiency and faster learning than a model-based algorithm. However, they did not do a systematic study investigating sample efficiency with different values of $\tau$ and their effect on the algorithm's sensitivity to other hyperparameters. We do this in our work. 

Varying $\tau$ can have interesting effects on DQN and its other hyperparameters. We investigate this hypothesis empirically in the Mountain Car environment~\cite{Moore1990}. Our objectives are to investigate whether (1) increasing $\tau$ helps DQN learn faster and achieve better performance, (2) different $\tau$ have different effects on sample efficiency, and (3) increasing $\tau$ makes DQN less sensitive to other hyperparameters, thereby making it easier to choose their best values. %In this draft, we provide preliminary results on DQN which confirm our hypothesis. To our best knowledge, no previous work has studied this phenomenon for deep RL algorithms.

\section{Background}
\label{background}

%***Background section should define the problem setting formally
%and any notation you will need later in the paper
%• I should never come across a symbol later in the paper that was
%not defined***

A Markov Decision Process (MDP) is defined by the tuple $(\mathcal{S}, \mathcal{A}, R, P, \gamma)$, where $\mathcal{S}$ and $\mathcal{A}$ denote the state and action space respectively. $R: \mathcal{S} \times \mathcal{A} \rightarrow \mathbb{R}$ denotes the scalar reward function, $P: \mathcal{S} \times \mathcal{S} \times \mathcal{A} \rightarrow [0, 1]$ is the state transition probability, where $P(s^\prime | s, a)$ denotes probability of transitioning to state $s^\prime$ from state $s$ by taking action $a$. $\gamma \in [0, 1)$ denotes the discount factor. The goal of the agent is to find a policy $\pi: \mathcal{A} \times \mathcal{S} \rightarrow [0, 1]$ $(\pi(a|s)$ is the probability of taking action $a$ in state $s$) that maximizes the expected return
\begin{equation*}
    Q^\pi(s, a) \doteq \mathbb{E}_\pi \left[ r_1 + \gamma r_2 + \dots | S_0 = s, A_0 = a \right],
\end{equation*}
%One of the ways to achieve this is to learn an estimate of the action value function $Q$ and act greedily with respect to it to generate a policy.
where $r_k$ is the random variable of reward at time step $k$. Q-learning is an algorithm to achieve this by directly learning the optimal action-value function $Q^*(s, a) \doteq \max_\pi Q^\pi(s, a)$ and deriving the policy from that ~\cite{Watkins:89}. The learning rule for Q-learning is given by
\begin{multline}\label{eq:q_learning}
    Q_{t+1}(s, a) = Q_{t}(s, a) + \alpha \left( Z^Q_t - Q_{t}(s, a) \right), \\
    Z^Q_t = r + \gamma \max_{a^\prime} Q_t(s^\prime, a^\prime),
\end{multline}
where $(s, a, r, s^\prime)$ denotes a transition from state $s$ by taking action $a$ to state $s^\prime$ with reward $r$. $Z^Q_t$ and $\alpha$ above denote the target and scalar learning rate respectively. Intuitively, Q-learning moves the Q-function estimates toward the target. Note that the subscript $t$ above denotes the time step at which an update is made, which can be different from the time step at which the transition is collected. 

Q-learning is an off-policy algorithm, meaning it learns about the greedy policy but collects data using a different policy. To encourage exploration, one way to collect data is according to an $\epsilon$-greedy policy that chooses a random action with probability $\epsilon$ and the greedy action with $1-\epsilon$.

The learning rule in (\ref{eq:q_learning}) updates the value of each $(s, a)$ pair without affecting the values of other pairs. Hence there is no generalization of value from one pair to another even if the pairs are similar. This makes it unrealistic to directly use (\ref{eq:q_learning}) for large state-action spaces. To address this issue, we need to turn to function approximation where we learn a parameterized Q-function with a fixed number of parameters and use it for all $(s, a)$ pairs. 

\subsection{Deep Q-Network}\label{bg:dqn}

Deep Q-Network (DQN) is a deep learning version of Q-learning ~\cite{Mnih2015}. Here the Q-function is a neural network parameterized by $\theta$, where $Q(s, a; \theta_t)$ is obtained by passing the state $s$ into a network with parameters $\theta_t$ and one output for each action $a$. The DQN stochastic gradient descent update to $\theta_t$ is given by
\begin{multline}\label{eq:dqn}
    \theta_{t+1} = \theta_t + \alpha \left( Z^{\text{DQN}}_t - Q(s, a; \theta_t) \right) \nabla_\theta Q(s, a; \theta_t), \\
    Z^{\text{DQN}}_t = r + \max_{a^\prime} Q(s^\prime, a^\prime; \theta^{-}).
\end{multline}
One difference from Q-learning here is that the target uses $\theta^-$, which are the parameters of the target Q-network. The target Q-network is a lagging copy of the online Q-network, which is refreshed after every $C$ steps, i.e., $\theta^- = \theta_t$ at time step $t$, and then kept constant for the next $C$ steps. Using a target network induces stability when learning with neural networks by creating a delay between the time target is computed and the time when parameters are updated. 

Another important component and difference of DQN from standard Q-learning is experience replay, which we discuss in more detail in the following section. 

\subsection{Experience Replay}\label{bg:er}

Experience Replay (ER) was introduced by ~\cite{Lin1992}. It is a constant-size buffer with capacity $\Omega$ and comprising the agent's experiences (transition tuples) at interaction time steps, i.e. $e_k = (s_k, a_k, r_{k+1}, s_{k+1})$ at time step $k$. Note that a single transition tuple is sufficient to make updates according to (\ref{eq:dqn}). However, it is generally more effective to (randomly) sample a mini-batch of transitions (batch size $B$) from the buffer and use it to update Q-network parameters. 

Using an ER buffer provides multiple benefits~\cite{Mnih2015}. First, it improves sample efficiency by using a particular transition (sample) for multiple updates. Second and more importantly, it makes training more stable by randomizing the samples, thereby breaking correlations between consecutive samples.

A typical implementation of the replay buffer involves storing transitions at each time step and making room for new transitions by removing the old ones. This fixes the amount of time each transition spends in the buffer and helps to discard transitions from a very old policy, that might not be relevant to make the current update. The simplest strategy to sample transitions is sampling uniformly, i.e. each transition has an equal probability of being sampled. However, this does not efficiently use transitions that might be the most effective for training. Other sophisticated approaches like Prioritized ER address this issue by sampling important transitions more frequently ~\cite{Schaul2016}. While prioritization may provide a better performance, in this work we adhere to focusing on uniform random sampling. We do so to observe the sole effect of increasing the replay frequency $\tau$ in a simpler setting while keeping other components of the algorithm intact.

\comment{
\subsection{Double Q-learning and Double DQN}\label{bg:double_q_ddqn}

When observed carefully, the learning updates for standard Q-learning (\ref{eq:q_learning}) and DQN (\ref{eq:dqn}) use the same Q-function to determine the maximizing action and to evaluate it. Since our estimate of $Q$ is not perfect (at least early in learning), this can result in optimistically estimating the values of some actions in a state. This problem is formally referred to as maximization bias and can result in the degradation of the agent's performance~\cite{DoubleQ}. 

Double Q-learning provides a solution to this bias ~\cite{DoubleQ}. It learns two separate Q-functions and uses one for action selection and the other for evaluation. Doing so reduces the bias since a separate $Q$ is used for evaluating the maximizing action. Formally, if $Q^1$ and $Q^2$ are the estimates of the Q-function, the update to $Q^1$ is given by
\begin{multline}\label{eq:double_q_learning}
    Q^1_{t+1}(s, a) = Q^1_{t}(s, a) + \alpha \left( Z^{\text{DoubleQ}}_t - Q^1_{t}(s, a) \right), \\
    Z^{\text{DoubleQ}}_t = r + \gamma Q^2_t \left(s^\prime, \arg\max_{a^\prime} Q^1_t(s^\prime, a^\prime) \right).
\end{multline}
The update to $Q^2$ is symmetric with $Q^1$ and $Q^2$ reversed. Note that at each time step, we pick one of the Q-functions uniformly and update it. Hence the computation time per step is the same as Q-learning. 

With the above modification, it is intuitive to extend the idea of double Q-learning to DQN. The Double DQN (DDQN) algorithm does this ~\cite{hasselt2015doubledqn}. However, instead of maintaining a second Q-network, DDQN uses the target network to estimate the value of the maximizing action (which is selected using the online network). Hence the update is given by
\begin{multline}\label{eq:ddqn}
    \theta_{t+1} = \theta_t + \alpha \left( Z^{\text{DDQN}}_t - Q(s, a; \theta_t) \right) \nabla_\theta Q(s, a; \theta_t), \\
    Z^{\text{DDQN}}_t = r + Q \left(s^\prime, \arg\max_{a^\prime} Q(s^\prime, a^\prime; \theta_t) ; \theta^- \right).
\end{multline}
Note that the target network is not completely disjoint from the online network. However, using it for action evaluation gives good performance and saves computation by avoiding training another Q-network ~\cite{hasselt2015doubledqn}. 

Finally, we combine all the above ideas in Algorithm \ref{alg:dqn_ddqb_er} that illustrates DQN and DDQN under varying amounts of replay per step (replay frequency $=\tau$). 
\comment{
\begin{algorithm}[tb]
   \caption{DQN and DDQN under varying amounts of replay per step}
   \label{alg:dqn_ddqb_er}
\begin{algorithmic}[1]
   \STATE {\bfseries Initialize:} replay memory $D$ to capacity $B$ %and replay frequency $\tau$
   \STATE {\bfseries Initialize:} main Q-network with parameters $\theta$
   \STATE {\bfseries Initialize:} target Q-network with parameters $\theta^-$
   \STATE {\bfseries Input:} $\epsilon, \tau, C$
   \FOR{$\text{episode\footnotemark}=1$ {\bfseries to} $M$}
        \STATE Initialize the start state $s_0$
        \FOR{$k=0$ {\bfseries to} $T-1$}
            \STATE With probability $\epsilon$, select a random action $a_k$
            \STATE Otherwise select $a_k = \arg\max_a Q(s_k, a; \theta)$
            \STATE Execute action $a_k$ and observe reward $r_{k+1}$ and next state $s_{k+1}$
            \STATE Store transition $(s_k, a_k, r_{k+1}, s_{k+1})$ in $D$
            \FOR{$\text{replay}=1$ {\bfseries to} $\tau$}
                \STATE Sample a random mini-batch of transitions $(s_j, a_j, r_{j+1}, s_{j+1})$ uniformly
                \STATE 
                $a^\prime=\begin{cases}
                \arg\max_u Q(s_{j+1}, u; \theta^-) & \text{for DQN}\\
                \arg\max_u Q(s_{j+1}, u; \theta) & \text{for DDQN}
                \end{cases}$
                \STATE 
                $Z_j=\begin{cases}
                r_{j+1} & \text{for terminal $s_{j+1}$} \\
                r_{j+1} + Q(s_{j+1}, a^\prime; \theta^-) & \text{otherwise}
                \end{cases}$
                \STATE Perform a gradient descent step on $\left( Z_j - Q(s_j, a_j; \theta) \right)^2$ with respect to $\theta$
            \ENDFOR
            \STATE Every $C$ steps, refresh target network $\theta^- = \theta$
        \ENDFOR
   \ENDFOR
\end{algorithmic}
\end{algorithm}
\footnotetext{We use episodes just for clarity in the algorithmic formulation. For experiments, we instead fix the total number of environment interaction steps for an agent.}
}

}
\section{Experimental Design}\label{exp_design}
\comment{
In this section, we describe the Mountain Car environment that we use as a testbed for experiments. Moreover, we also provide the details of our experimental setup --how we have done our experiments, what hyper-parameters value we have used, and other required information for reproducing our results.}

In this section, we describe our environment setup, hyperparameter choices, and other experimental details in order to understand and reproduce our results. 

\subsection{Environment setup}
We evaluate DQN in the Mountain Car environment ~\cite{Moore1990}. Our experiments are based on OpenAI Gym's implementation of Mountain Car with slight modifications ~\cite{brockman2016openai}. The environment is formulated as an MDP having a two-dimensional continuous state space: position $\in [-1.2, 0.6]$ and velocity $\in [-0.07, 0.07]$. Note that this is the state of the agent and not the environment. The action space consists of three discrete actions: accelerate left $(0)$, do not accelerate $(1)$, and accelerate right $(2)$. The goal of the agent (car) is to reach the top of the hill as soon as possible. However, it does not have enough power to accelerate up the hill and hence should accelerate backward to generate enough momentum to climb up. A reward of $-1$ per step is given to encourage the agent to finish the task fast. The default AIGym implementation terminates an episode after $200$ steps. We modify this to $2000$ steps to avoid misleading results due to aggressive episode cutoffs~\cite{Patterson2020}. At the same time, we do not set it to infinity to avoid the agent getting stuck. 

\subsection{Experimental Setup}

Our objectives are to (1) assess the effect of increasing $\tau$ on the performance and sample efficiency of DQN in Mountain Car, and (2) understand the variety of behaviors produced by DQN with different $\tau$ values for different hyperparameter settings. For our experiments, we borrow terminology from ~\cite{Patterson2020} where an \textit{agent} refers to ``a single entity learning and adapting over time", an \textit{algorithm} refers to ``a process producing a set of agents by specifying initial conditions and learning rules", and \textit{hyperparameters} refer to ``a scalar set of parameters affecting the agents produced by the algorithm".

We fix the total number of environment interactions (steps) for each agent instead of episodes. Doing so ensures a fair evaluation as each agent receives the same amount of learning experience. We found $N=250,000$ steps good to evaluate if an agent reaches good performance and stably maintains it for some time. 

We use a discount factor $\gamma=0.99$ for all our experiments. However, we use the undiscounted return as our performance metric since discounting is a part of the agent's internal mechanics and not the environment. Using the undiscounted return for Mountain car indicates how quickly the task is solved. Further, we measure online performance, i.e. how the agent performs while it is learning. Note that it needs to balance exploration and exploitation in such evaluation ~\cite{Patterson2020}. To measure performance at a particular step of learning, we use the undiscounted return for the episode containing that step. 

We do 30 runs for each $\tau$, and each run corresponds to $250,000$ steps but differs in terms of the random seed (Q-network initialization and initial state of the agent). To fairly compare run $i$ of two agents with different $\tau$, we need to ensure that their seeds for run $i$ are the same. We do so in the following way:
\begin{itemize}
    \item For the Q-network initialization, we randomly generate an array of seeds of length equal to the total number of runs and use the same array for each $\tau$.
    \item For the initial state of the agent, the AIGym environment gets reinitialized every time a new episode starts inside a run. Hence we generate a large array of seeds to ensure that the same seed (initial state) is used for episode $j$ inside run $i$ for agents with different $\tau$.
\end{itemize}
Additionally, we use Xavier initialization to initialize the Q-network parameters ~\cite{xavier}. 

To achieve our first objective, we follow a simple strategy to set the hyperparameters for our experiments. We borrow most hyperparameter values of DQN in Mountain Car from an existing codebase\footnote{\href{https://github.com/andnp/rl-control-template/blob/master/experiments/example/MountainCar/DQN.json}{\textcolor{blue}{Link}} to the repository from which we took hyperparameters.} and set the remaining ones using random search. We use these hyperparameters for all values of $\tau$. The resulting setting of hyperparameters might not be the best but is good enough to ensure a nearly-steady improvement in performance. We argue that doing this is appropriate since we want to assess how much better DQN can be made just by increasing $\tau$ for a system that might not be exhaustively tuned for the best hyperparameters. 

Our main entity of interest is $\tau$ and we assess the change in performance upon increasing $\tau$ while keeping other hyperparameters fixed. We evaluate on $\tau=[1,2,4,8,16,32]$, where $\tau=1$ corresponds to vanilla DQN. The maximum number of transitions that can be stored in the replay buffer (capacity $\Omega$) is set to $4000$ and the number of transitions sampled per update (batch size $B$) is set to $32$. Further, at the beginning of each run, the initial policy collects and stores transitions for the first $1024$ steps (replay start size) without making any updates to the Q-network. We set the replay start size to be much bigger than the batch size to ensure better randomization while sampling that breaks correlations between samples in early learning. If the replay start size is smaller, there is a higher probability to sample the most correlated recent transitions, which can result in a bad initial start and never recovering thereafter. 

To represent the Q-network, we use a neural network with two hidden layers of size $32$ each and with the ReLU activation function. We refresh the parameters of the target Q-network every $C=128$ steps and use a mean-squared loss to measure the difference between the DQN target and Q value. To optimize this loss function, we use the Adam optimizer with learning rate $\alpha=0.001$, gradient momentum $\beta_1=0.9$, and squared gradient momentum $\beta_2=0.999$. Finally, we use an $\epsilon$-greedy behavioral policy for exploration with $\epsilon_{initial}=1$ at the start of a run, decayed to $\epsilon_{final}=0.1$ with a decay rate $\epsilon_{decay}=0.999$ and fixed thereafter. This roughly corresponds to annealing $\epsilon$ from $1$ to $0.1$ over $2300$ steps. 

For our second objective, we do hyperparameter sensitivity analysis. Sensitivity plots help us understand changes in the behavior of algorithms and suggest sensitivity to hyperparameters that is essential for deployment \cite{Patterson2020}. We pick $4$ hyperparameters: learning rate, batch size, replay capacity, and the target network refresh rate. We chose these hyperparameters as we found them to be amongst the most important ones affecting DQN's performance. The range of tested values is specified in the next section. 

%Clearly identify the problem setting: exactly what problem
%are you addressing and for what specific setting?

%specify environments and the experiments you plan to run (algorithms, baselines, evaluation scheme)

\section{Evaluation and Results}
\label{evaluation}

As stated before, we care about online performance, which is measured at each step and equals the undiscounted return for the episode containing that step~\cite{Patterson2020}. If an episode finishes with an undiscounted return of $G$, then every step of that episode has the same performance value, $G$. We refer to this as the performance measure.

\begin{figure}[ht]
\vskip -0.1in
\begin{center}
\centerline{\includegraphics[width=0.9\columnwidth]{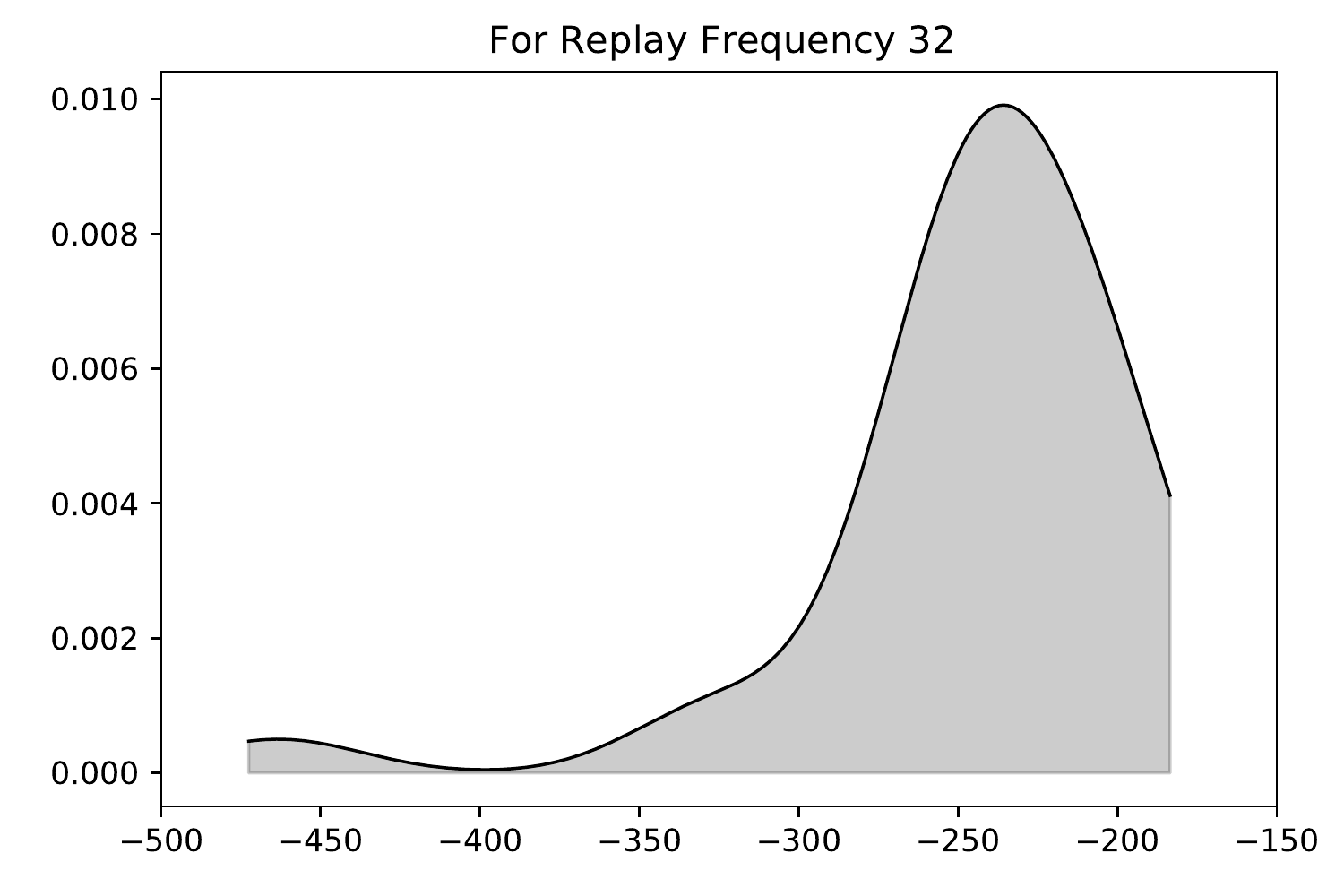}}
%\vspace*{-5mm}
\caption{Performance distribution $\mathbb{P}(M)$ with $\tau=32$ for $30$ runs.}
\label{fig:distribution_rf_32}
\end{center}
%\vskip -0.2in
\vspace*{-8mm}
\end{figure}

To aggregate performance for a single run, we simply sum the performance measures at each step. To get a scalar aggregate performance value over multiple runs for each replay frequency, we sum the aggregate performance for each run and divide it by the total number of steps times the total number of runs. Additionally, to get the mean performance curve, we average each step's performance measure over all the runs and do this for all steps. Note that we do not apply any kind of smoothing for the mean performance curve. For instance, it is possible to sum the performance measures till a particular step, divide the sum by the step index, and use that for plotting the mean curve (like a running average). In such a case, the curve becomes very smooth. However, we do not do this as it hides potentially important variations. 

Since individual runs can be different from each other depending on the seed initialization, we use the following techniques to measure the variability in performance~\cite{Patterson2020} along with the mean performance:
\begin{itemize}
    \item Confidence Interval: We use confidence intervals to measure the uncertainty in our estimated mean online performance. For computing it, we use the Student t-distribution, which requires the assumption of having an approximately Gaussian performance distribution. To validate this assumption in our case, we visualize the distribution over performance. We use a $95\%$ confidence interval ($\alpha = 0.05$) to report our uncertainty in our mean estimate.
    \item Tolerance Interval: We use tolerance intervals to capture the variation in the performance from a limited number of samples. We use a ($\alpha = 0.05, \beta = 0.9$) tolerance interval to examine the performance variability between multiple runs. The $(\alpha,\beta)$ values suggest that the interval contains at least $0.9$ fraction of total runs with the confidence of $95\%$.  
\end{itemize}

Now we provide the empirical results\footnote{We provide only the main empirical plots in this draft for brevity. To see all plots, please check \href{https://drive.google.com/drive/folders/1AmAuM-v8expJ6BuGCsWzPr2rurKOhm0S?usp=share_link}{\textcolor{blue}{this}} drive link.}. First, we visualize the distribution of performance for DQN. Fig.~\ref{fig:distribution_rf_32} shows the approximate distribution for $30$ runs with replay frequency $\tau=32$. This distribution is obtained by computing the aggregate performance for each run, dividing it by the total interactions, and plotting the frequency distribution. Hence a sample of this distribution corresponds to a run. Note that the resulting distribution is approximately Gaussian, which allows using the Student-t confidence intervals\footnote{We use t $= 2.045$ from the Student-t table for 30 runs.}. The x-axis contains the distribution between $-472.57$ and $-183.51$. This is because the aggregate performance for each run (sample) among the $30$ runs was between these two values. We do not show the performance distribution for other $\tau$ values because they follow a similar behavior with the difference that the distribution for larger $\tau$ values has a greater mean and lower variance estimate than smaller $\tau$ values. 

%This distribution is obtained by computing the undiscounted return for each episode across all runs and plotting the frequency distribution. 

\begin{figure*}[!t]
    \centering
    \begin{subfigure}{.33\textwidth}
        \centering
        \includegraphics[width=1\textwidth]{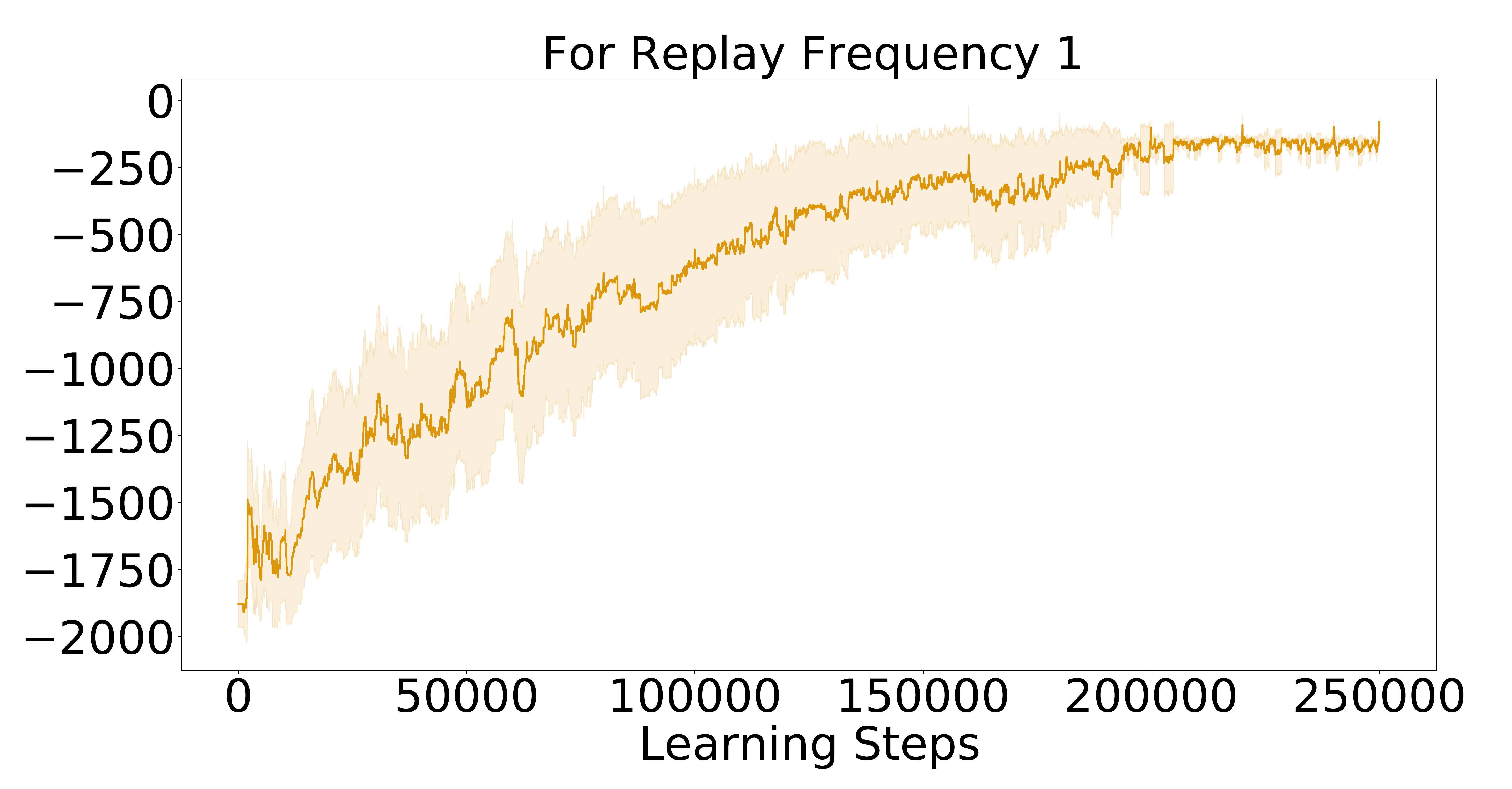}
        \subcaption{}
    \end{subfigure}%
    \begin{subfigure}{.33\textwidth}
        \centering
        \includegraphics[width=1\textwidth]{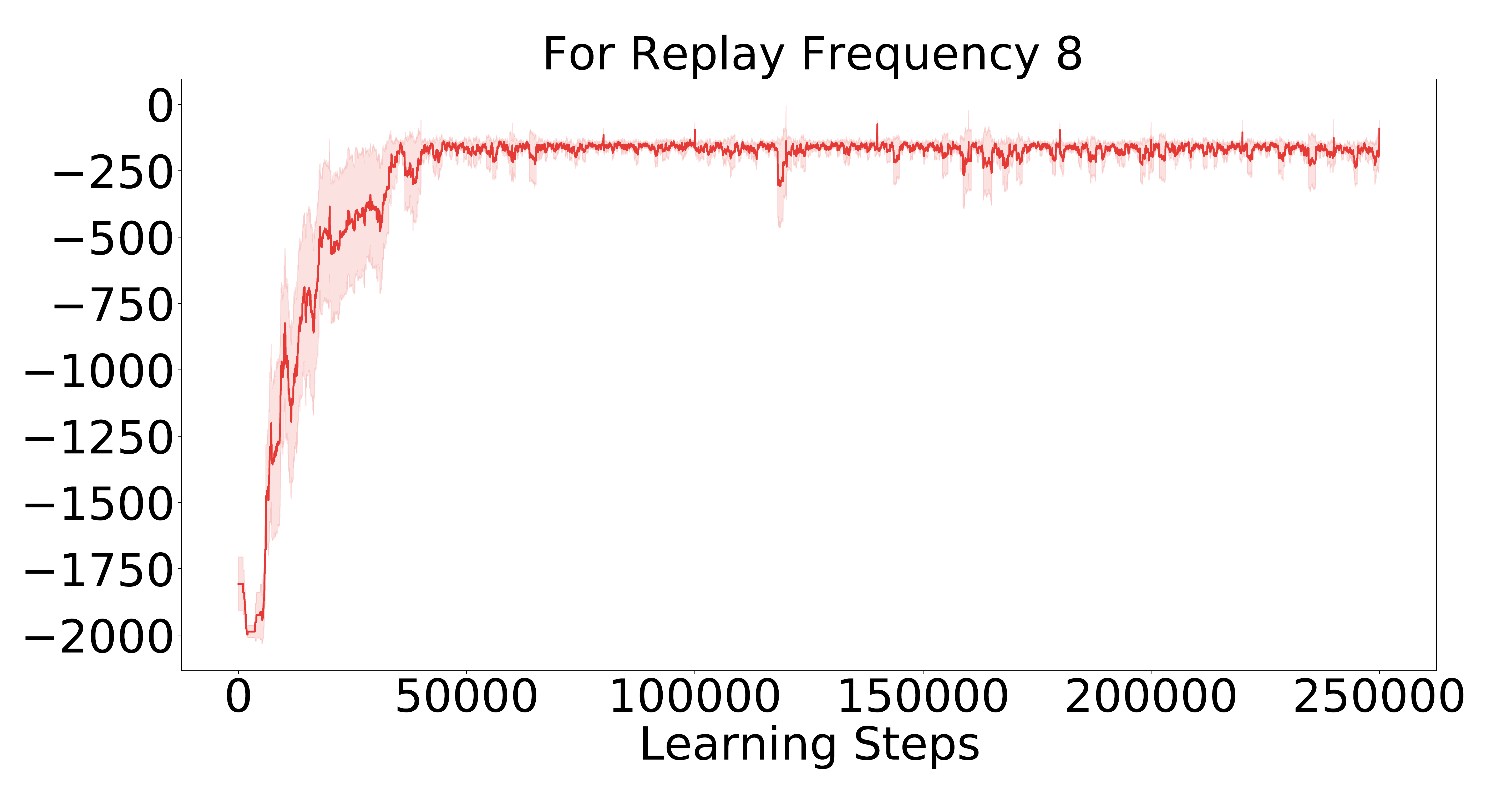}
        \subcaption{}
    \end{subfigure}%
    \begin{subfigure}{.33\textwidth}
        \centering
        \includegraphics[width=1\textwidth]{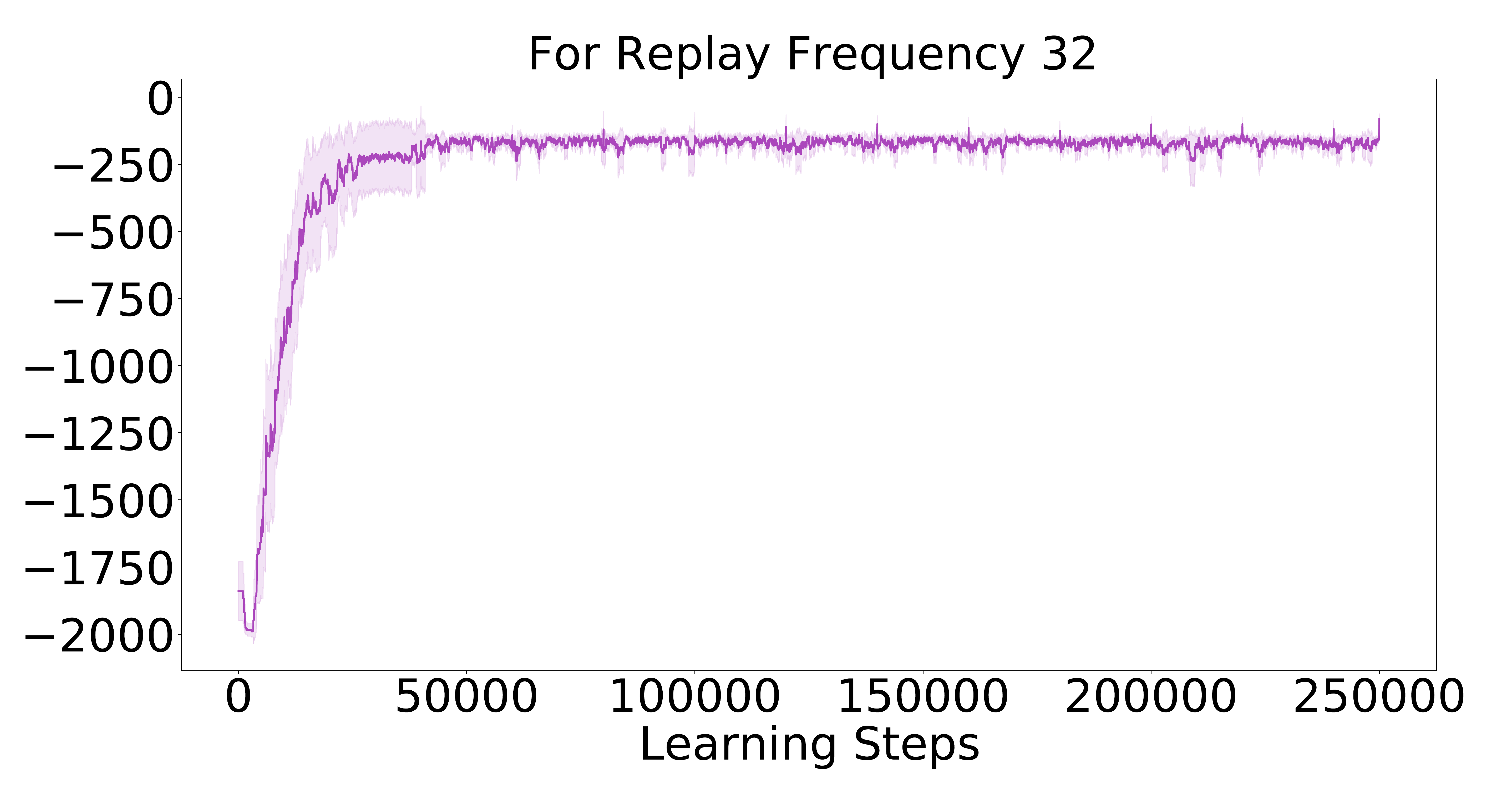}
        \subcaption{}
    \end{subfigure}

    \begin{subfigure}{.33\textwidth}
        \centering
        \includegraphics[width=1\textwidth]{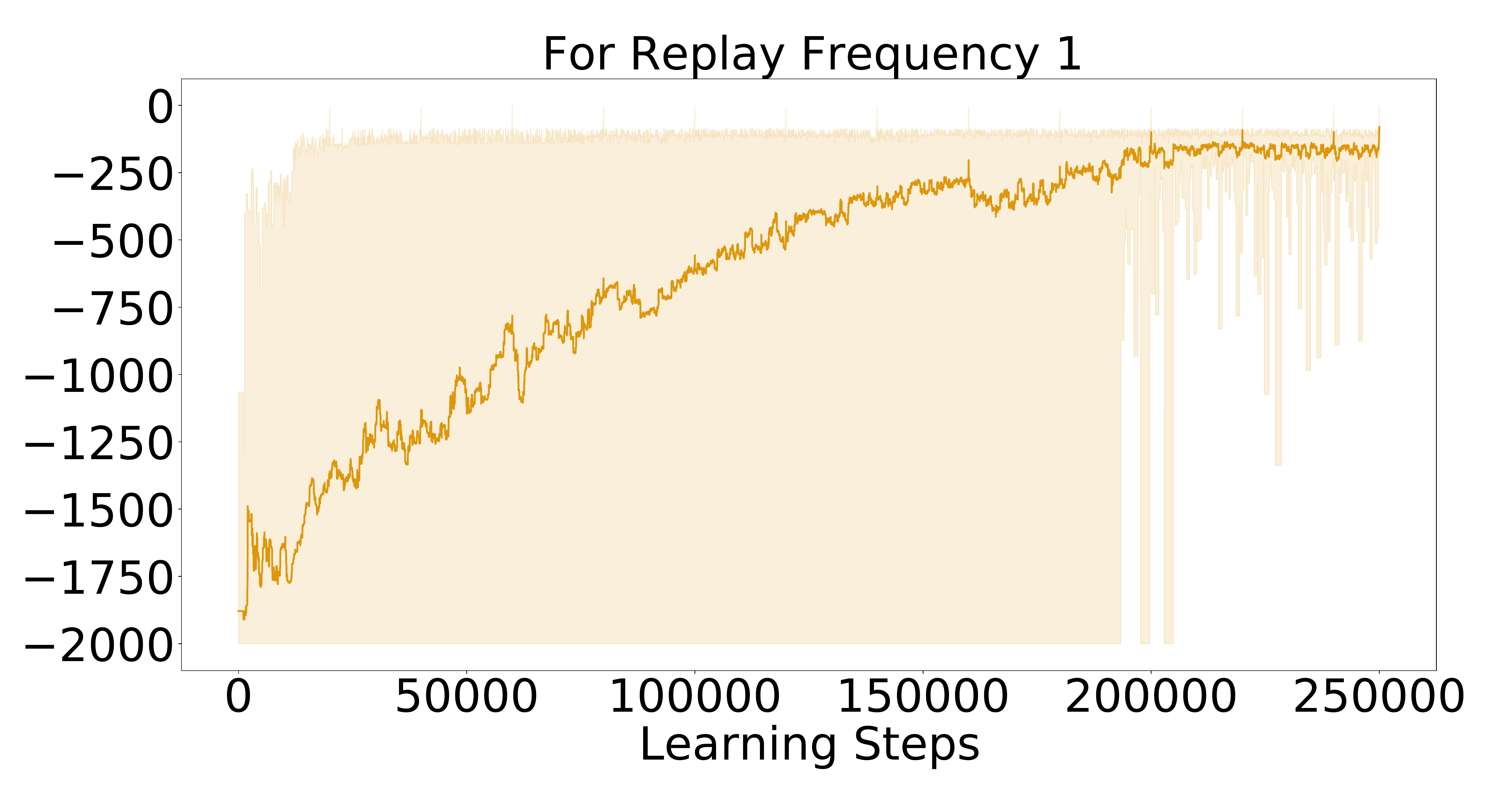}
        \subcaption{}
    \end{subfigure}%
    \begin{subfigure}{.33\textwidth}
        \centering
        \includegraphics[width=1\textwidth]{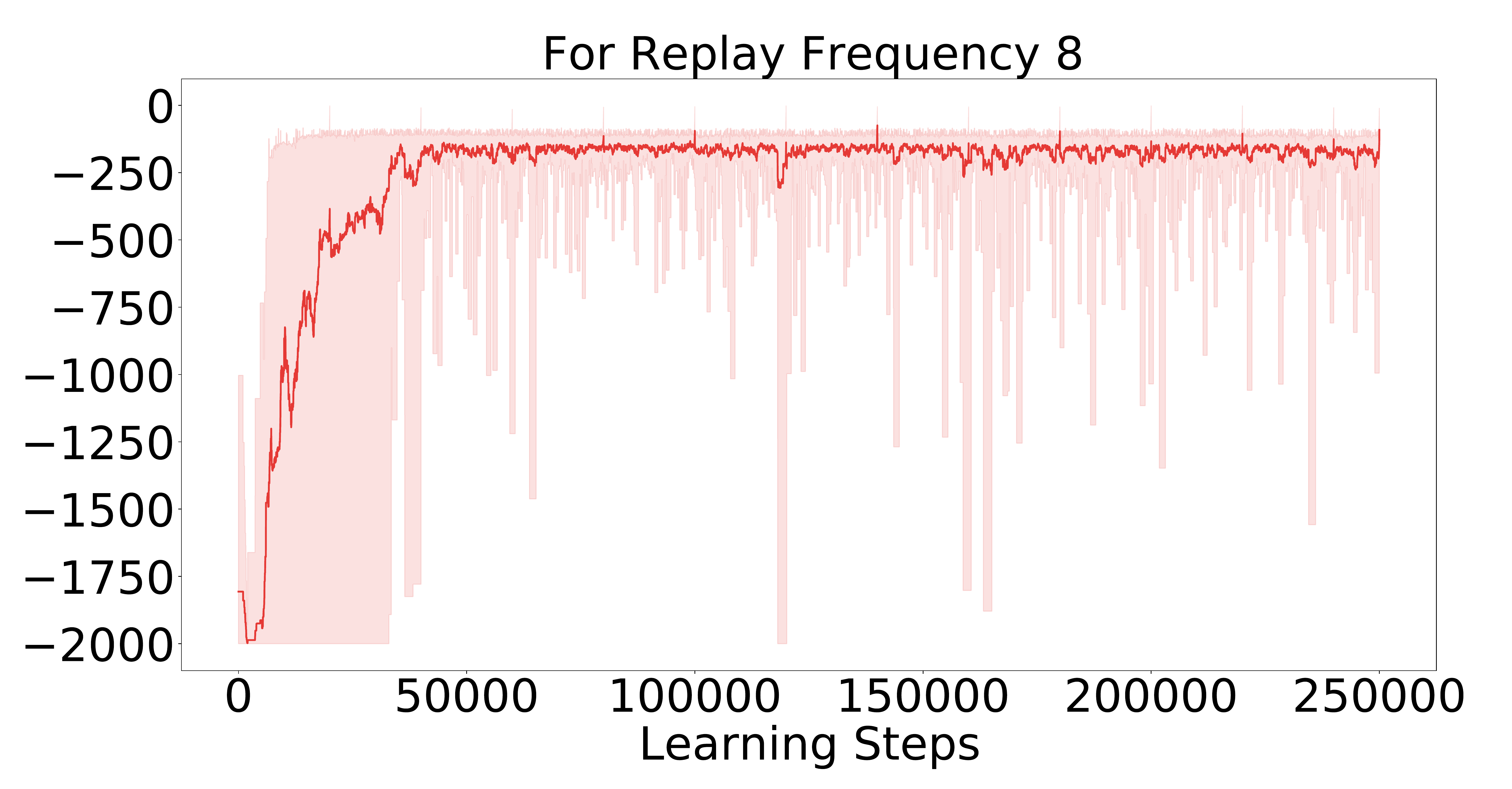}
        \subcaption{}
    \end{subfigure}%
    \begin{subfigure}{.33\textwidth}
        \centering
        \includegraphics[width=1\textwidth]{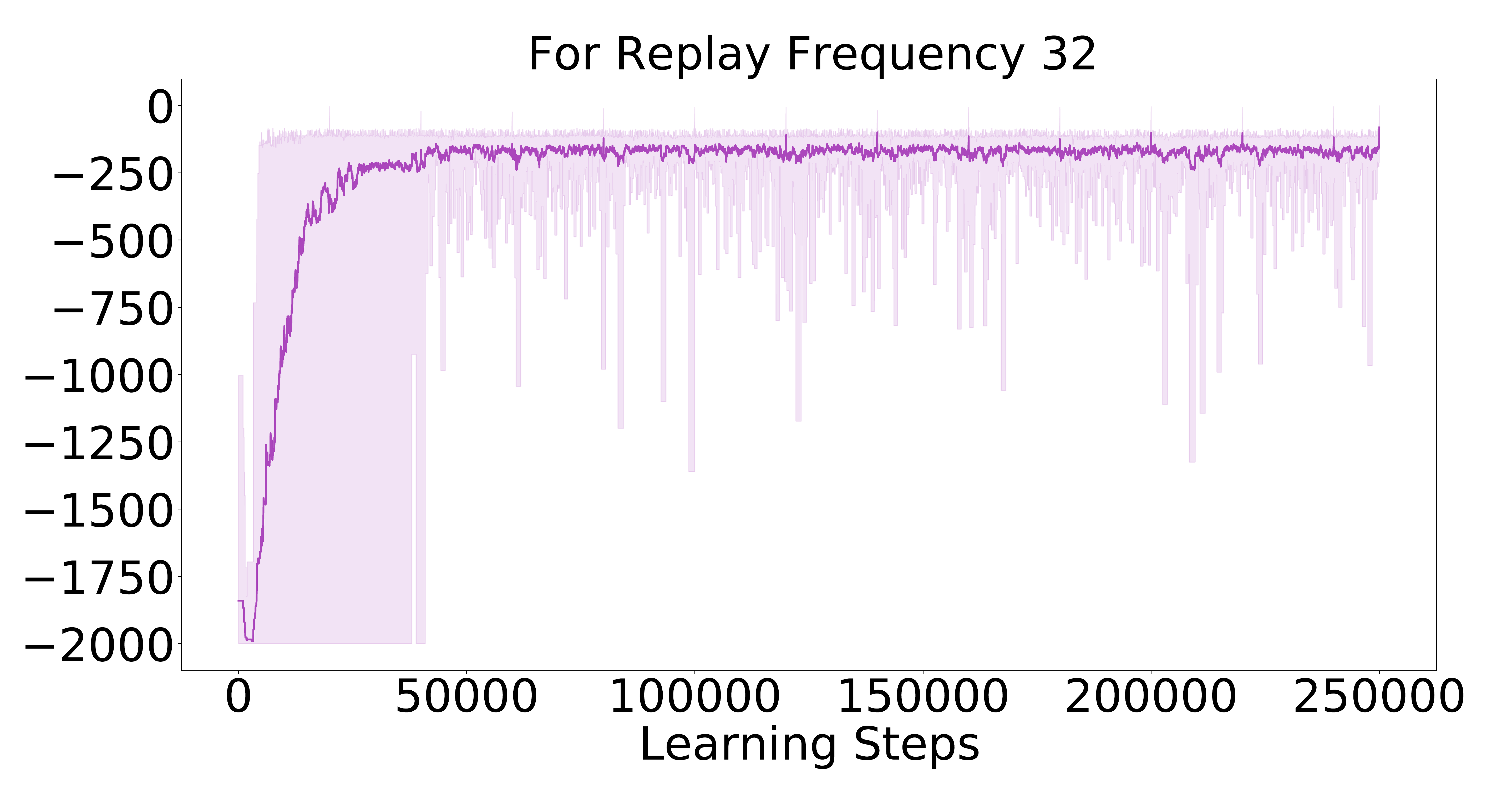}
        \subcaption{}
    \end{subfigure}
    
      \begin{subfigure}{.33\textwidth}
        \centering
        \includegraphics[width=1\textwidth]{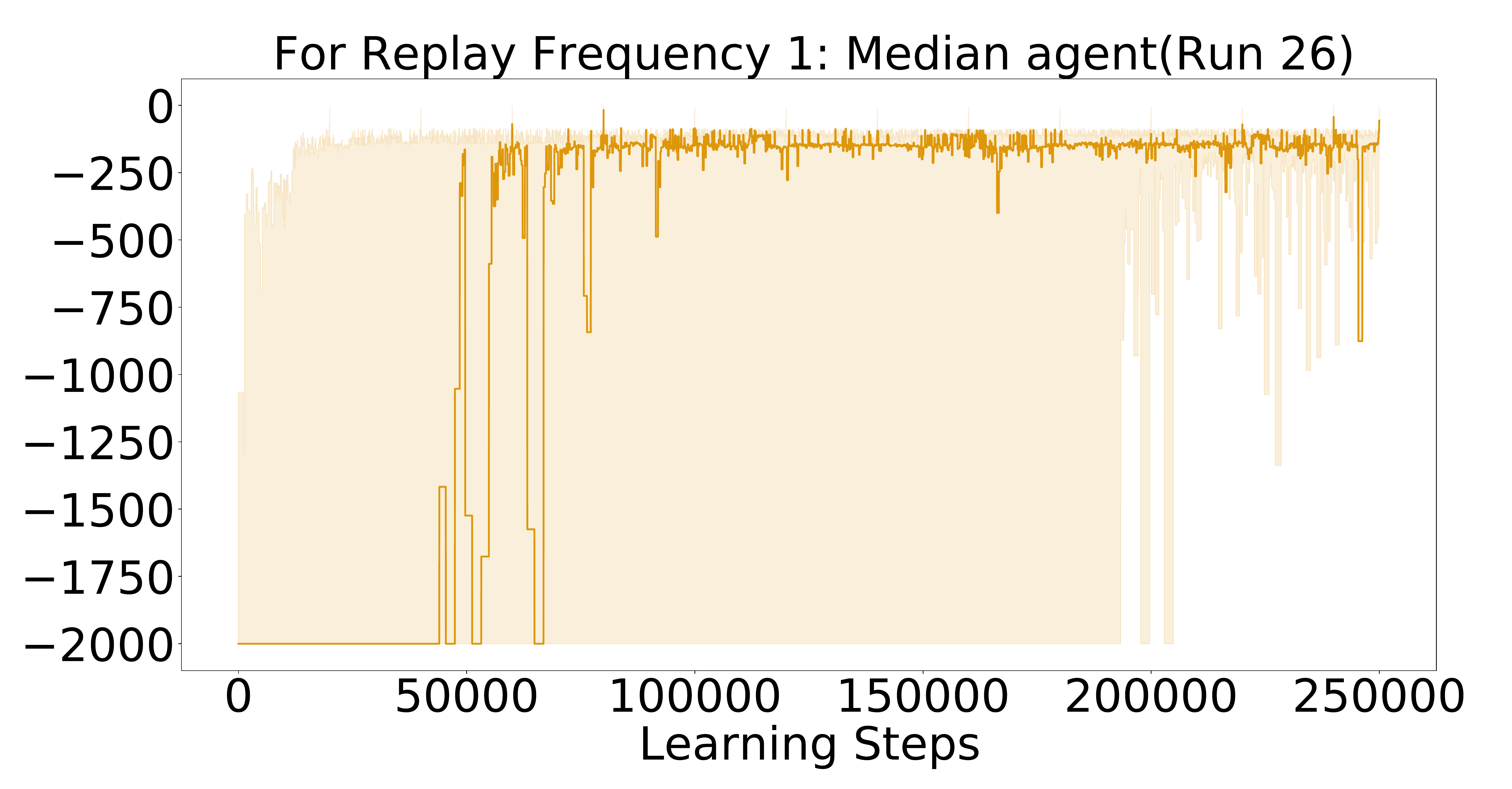}
        \subcaption{}
    \end{subfigure}%
    \begin{subfigure}{.33\textwidth}
        \centering
        \includegraphics[width=1\textwidth]{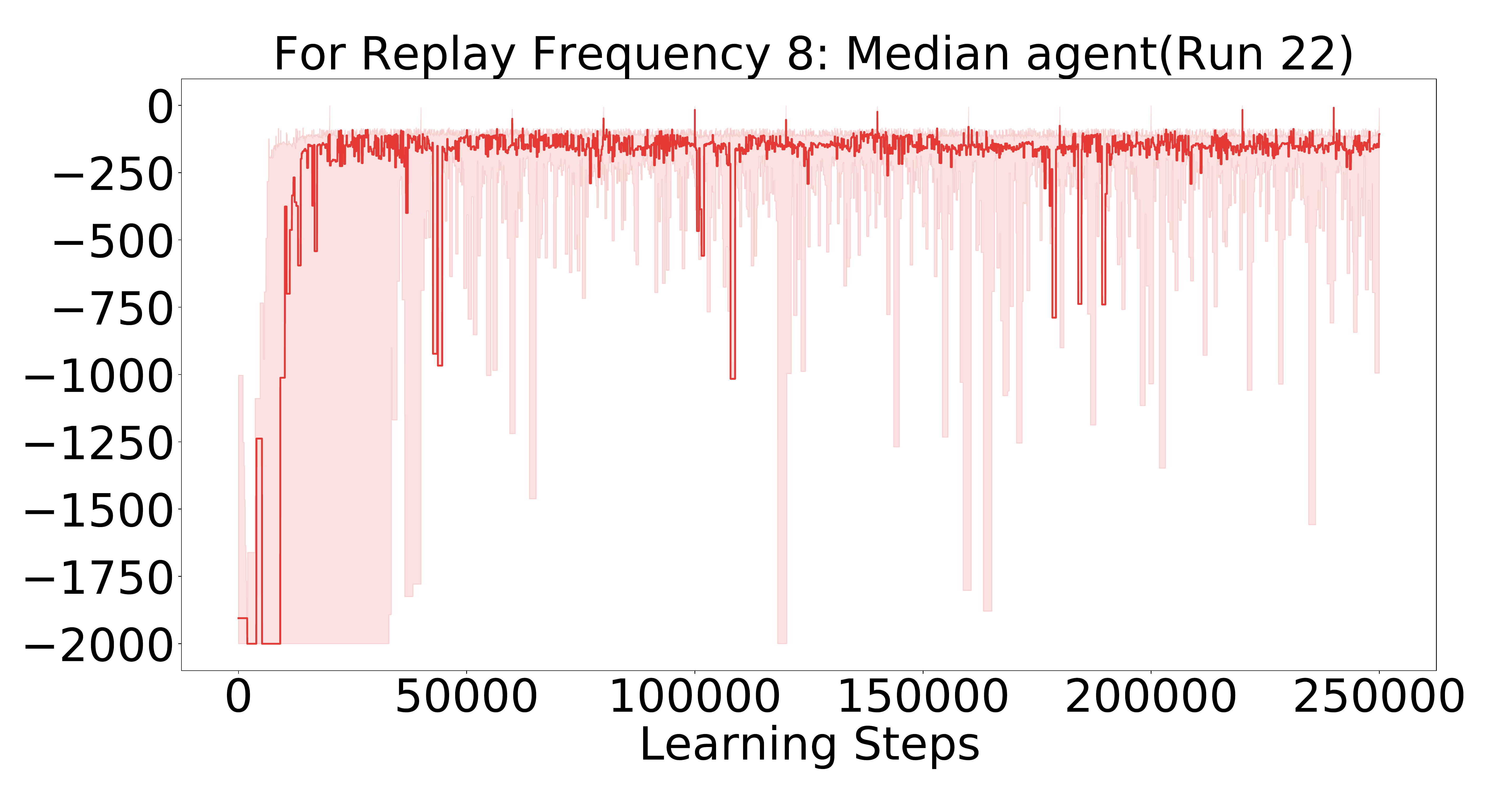}
        \subcaption{}
    \end{subfigure}%
    \begin{subfigure}{.33\textwidth}
        \centering
        \includegraphics[width=1\textwidth]{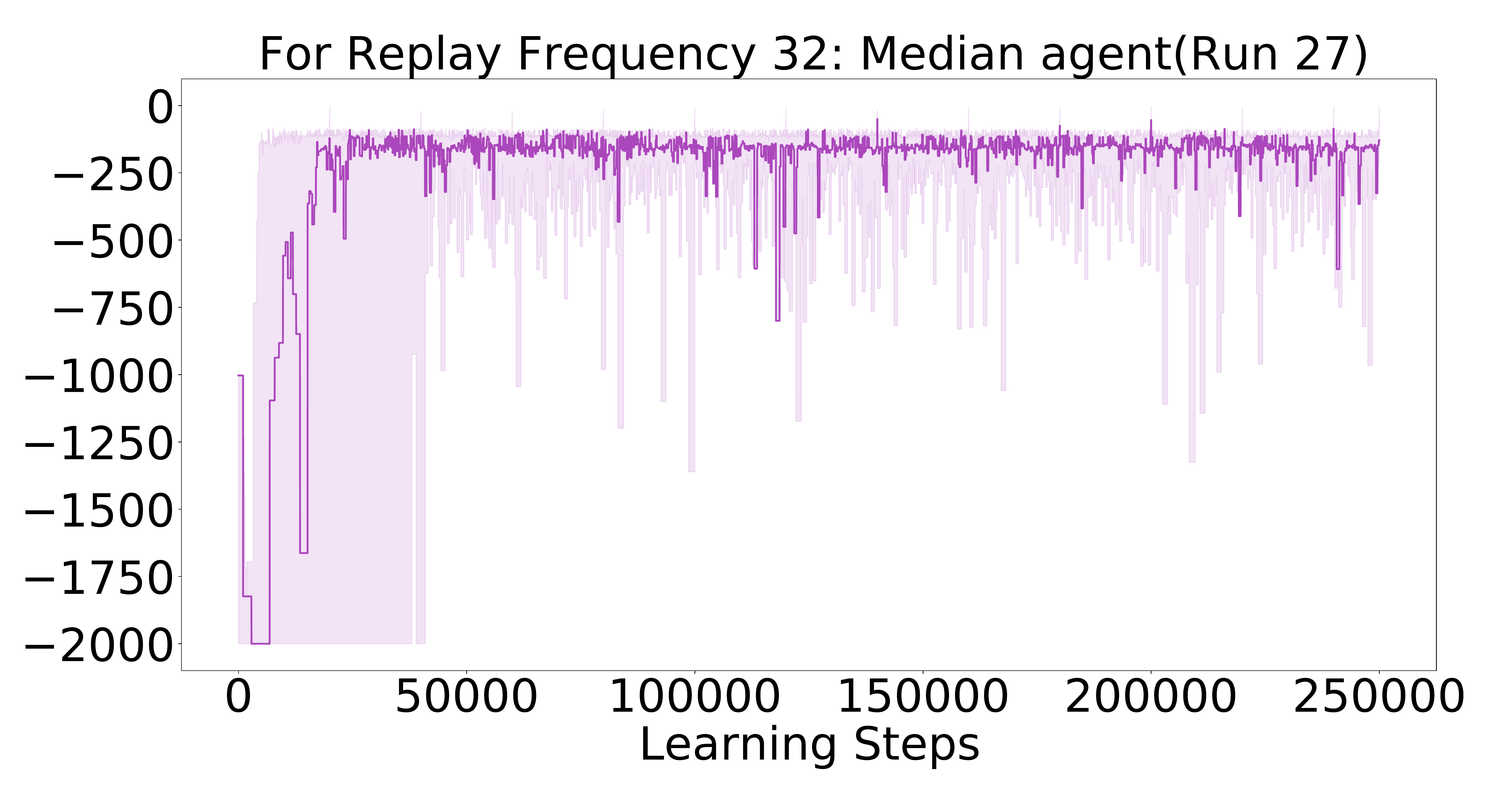}
        \subcaption{}
    \end{subfigure}
    
    \caption[short]{DQN performance in terms of confidence and tolerance intervals with different replay frequencies $\tau$: (a, b, c) show the confidence intervals with mean performance for $30$ runs and $\tau=(1, 8, 32)$ respectively. The curves (d, e, f) show the tolerance intervals with the mean performance for $30$ runs and $\tau=(1, 8, 32)$ respectively. The curves (g, h, i) show the tolerance intervals with the median agent's performance for $29$ runs and $\tau=(1, 8, 32)$ respectively. The y-axis denotes performance. No smoothing has been applied to the curves.}
    \label{fig:dqn_confidence_tolerance_intervals}
\end{figure*}

\subsection{Confidence Intervals}

The mean performances and confidence intervals for $\tau=(1, 8, 32)$ with $30$ runs are shown in fig.~\ref{fig:dqn_confidence_tolerance_intervals} (a, b, c), respectively. The interval for $\tau=1$ is wider than $\tau=8$, which is wider than $\tau=32$ (when we say $\tau=i$, we imply DQN with $\tau=i$). This shows that with a higher replay frequency, we are more certain in our estimate of the mean. %For $\tau=32$, the interval is slightly wider than $\tau=8$. This might be because fewer runs are used for $\tau=32$. We expect the interval to shrink if we increase the number of runs to $30$. 
The three plots also show that $\tau=32$ learns faster than $\tau=8$, which learns faster than $\tau=1$, i.e. uses fewer samples. Note that agents with all replay frequencies eventually achieve good performance. However, a higher $\tau$ results in better sample efficiency. This confirms that using more computation per step with fixed data (from the replay buffer) results in faster learning for DQN in Mountain Car. 

\comment{
Fig.~\ref{fig:dqn_confidence_interval_comparision} shows a comparison between the mean performances and confidence intervals for all $\tau$ values. We can see that the agent reaches near-optimal performance in a lesser number of steps for higher $\tau$ values. This shows that sample efficiency improves by increasing $\tau$. Moreover, the improvement is large and clearly visible from $\tau=1$ to $\tau=2$ and from $\tau=2$ to $\tau=4$. Note that agents with all replay frequencies ultimately achieve near-optimal performance in $250,000$ learning steps. However, we show only the first $100,000$ steps in fig.~\ref{fig:dqn_confidence_interval_comparision} to press on the observation that increasing replay frequency solves the task faster, i.e. uses fewer samples. }

Lastly, we make a few observations about stability with different $\tau$ values. From fig.~\ref{fig:dqn_confidence_tolerance_intervals} (a, b, c), we can observe that the mean performance is roughly stable for $\tau=(8,32)$ after about $50,000$ learning steps (the same holds true for $\tau=(4,16)$ but plots are not included for brevity). For $\tau=(1,2)$, the mean starts to stabilize after $200,000$ and $150,000$ steps respectively. This shows that with a higher $\tau$, not only does the agent reach good performance faster but also maintains that on average. The mean curves are still a little noisy because we do not apply any smoothing. 

\comment{
\begin{figure}[p]
\vskip 0.2in
\begin{center}
\centerline{\includegraphics[width=\columnwidth]{figs/confidence Interval_runs_30_for_rf_1_2_4_8_run_15_for_rf_16_32.pdf}}
\caption{Comparison of confidence intervals with different replay frequencies for DQN. These results are averaged over multiple runs, the x-axis denotes the learning steps, and the y-axis denotes the average performance measure over multiple runs. Note: Our experiment uses $250,000$ learning steps, but in this figure, we only show the first $100,000$ to improve readability. Please note that no smoothing is applied to the curves.}
\label{fig:dqn_confidence_interval_comparision}
\end{center}
\vskip -0.2in
\end{figure}
}

\subsection{Tolerance Intervals}

Tolerance intervals summarize the range of an algorithm's performance, irrespective of the underlying performance distribution while taking into account the uncertainty due to a limited number of samples. To compute tolerance intervals, we use the method described in~\cite{Patterson2020}. Fig.~\ref{fig:dqn_confidence_tolerance_intervals} (d-i) depict the tolerance intervals around the mean performance and around the median agent's performance. 

Fig.~\ref{fig:dqn_confidence_tolerance_intervals} (d, e, f) show the interval around the mean performance for $\tau=(1, 8, 32)$, respectively. Note that the interval is much wider for $\tau=1$ than for $\tau=8$. Further, the interval for $\tau=32$ is tighter than for $\tau=8$. This shows that with a higher $\tau$, the variation in algorithm performance is low. %The interval for $\tau=32$ is tighter than $\tau=8$ even though the former uses just $15$ runs. We expect that with more runs, the interval for $\tau=32$ will approach the true variation in performance. 
The tolerance intervals also show the bottom percentile of runs which indicates that the worst-case performance of a higher $\tau$ ($\tau=8, 32$) is better than a lower $\tau$ ($\tau=1$).

Fig.~\ref{fig:dqn_confidence_tolerance_intervals} (g, h, i) show the interval around the median agent's performance. The learning curve for the median agent is obtained by arranging the aggregate performances for the first $29$ runs\footnote{Having an even number of runs ($30$) requires averaging the middle two runs after arranging the aggregated runs in increasing order. However, the average is not representative of any single run and can hide the differences between the behavior of the individual runs. Hence we use $29$ runs for the median.} in increasing order, finding the median, and plotting the curve for the corresponding run index. Note that the learning curve for a higher $\tau$ even for an individual run (median here) is not very noisy and the performance increases and stays between $-100$ and $-200$ most of the time except for a few occasional drops (verified empirically). 

\subsection{Replay Frequency Curve}

\begin{figure}[t]
\vskip -0.05in
\begin{center}
\centerline{\includegraphics[width=0.9\columnwidth]{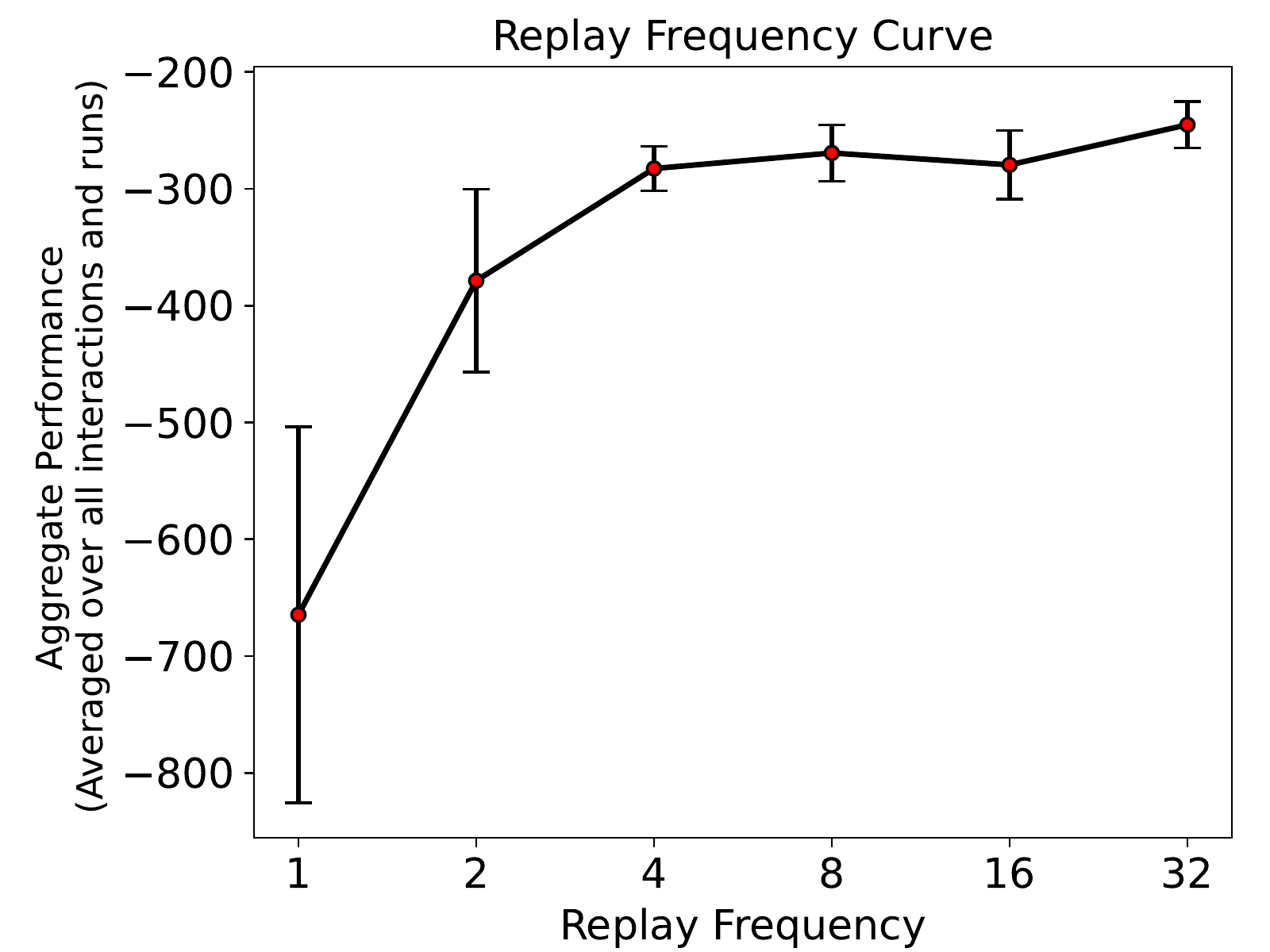}}
\caption{Comparison of different replay frequencies in terms of aggregate performance measure and error bar (computed using Student t-distribution) over $250,000$ interactions and $30$ runs for DQN. The x-axis is on the log scale (base $2$).\\}
\label{fig:dqn_replay_curve}
\end{center}
\vskip -0.6in
\end{figure}

To get a bigger picture of DQN performance with increasing replay frequency, we plot the aggregate performance against replay frequencies in fig.~\ref{fig:dqn_replay_curve}. The y-values denote the aggregate (online) performance across all $30$ runs each with $250,000$ steps, which is computed using the method described in the second paragraph of section~\ref{evaluation}. The error bars are computed using Student-t confidence intervals and depict the uncertainty in the mean estimates. The curve indicates how well DQN with different replay frequencies performs, given a fixed number of interactions with the environment. Hence greater y-values denote better sample efficiency. Note that the curve does not depict the final policy learned after training. The performance of the final policy is better than the average aggregate performance during training shown in the curve.

As shown in fig.~\ref{fig:dqn_replay_curve}, vanilla DQN ($\tau=1$) performs the worst. When the replay frequency is increased to $\tau=2$ and $\tau=4$, the increase in mean performance estimate is large and the uncertainty reduces. However, after $\tau=4$, the change is not very large. Interestingly, the mean and uncertainty estimates degrade slightly when moving from $\tau=8$ to $\tau=16$ but improve from $\tau=16$ to $\tau=32$. From $\tau=4$ onward, the confidence varies but is still more than $\tau=(1, 2)$.  

Even though aggregation hides the internal behavior of individual runs, fig.~\ref{fig:dqn_replay_curve} helps find the suitable replay frequency for a given scenario. For instance, if we have computational constraints, we would prefer using $\tau=4$ because it provides a good enough performance without a large increase in computation per step. However, if we care more about performance, we might trade $8$ times more computation per step for an improvement in performance. 

\comment{
\begin{figure*}[!t]
    \centering
    \begin{subfigure}{.23\textwidth}
        \centering
        \includegraphics[width=1\textwidth]{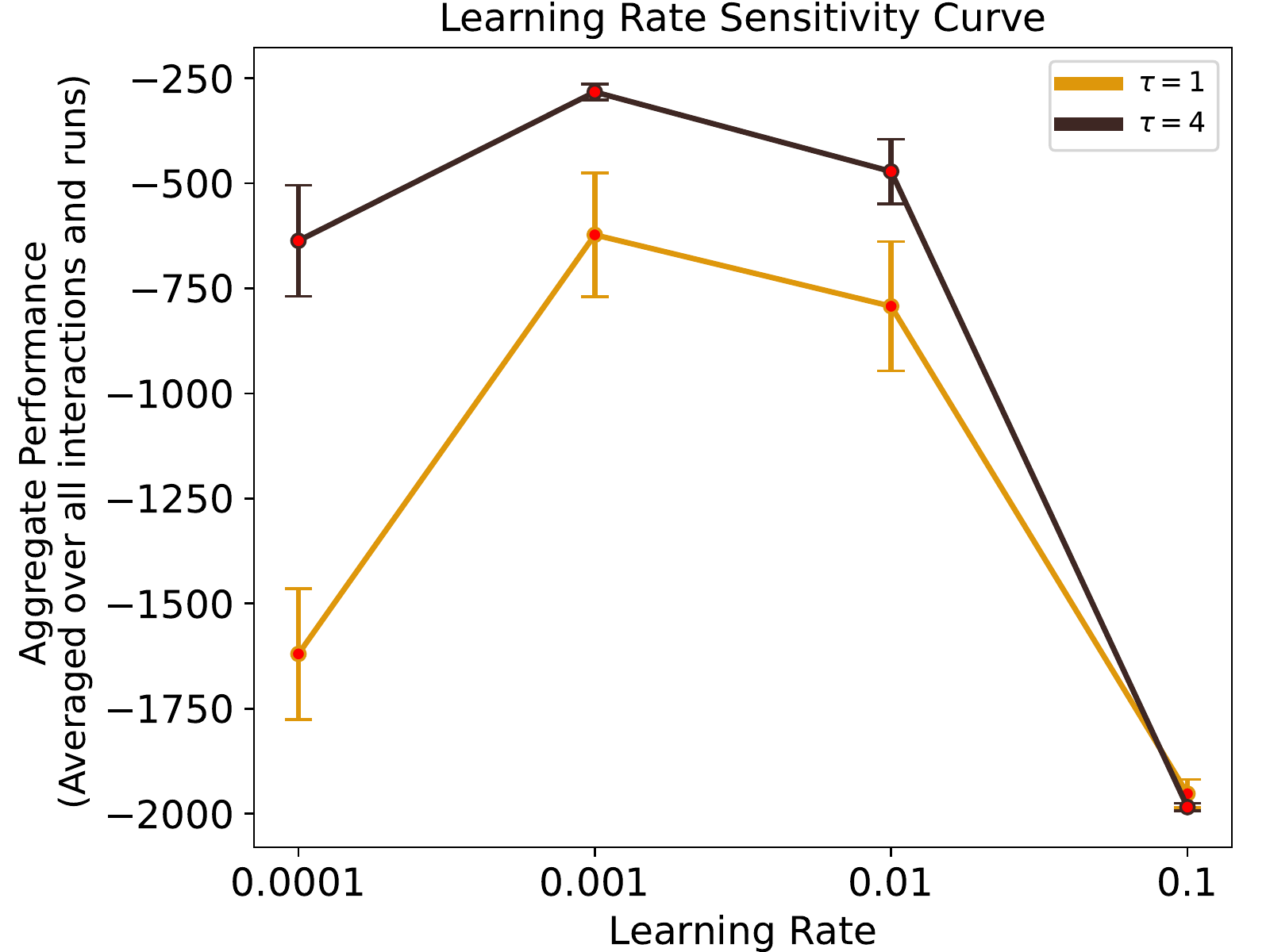}
        \subcaption{}
    \end{subfigure}%
    \begin{subfigure}{.23\textwidth}
        \centering
        \includegraphics[width=1\textwidth]{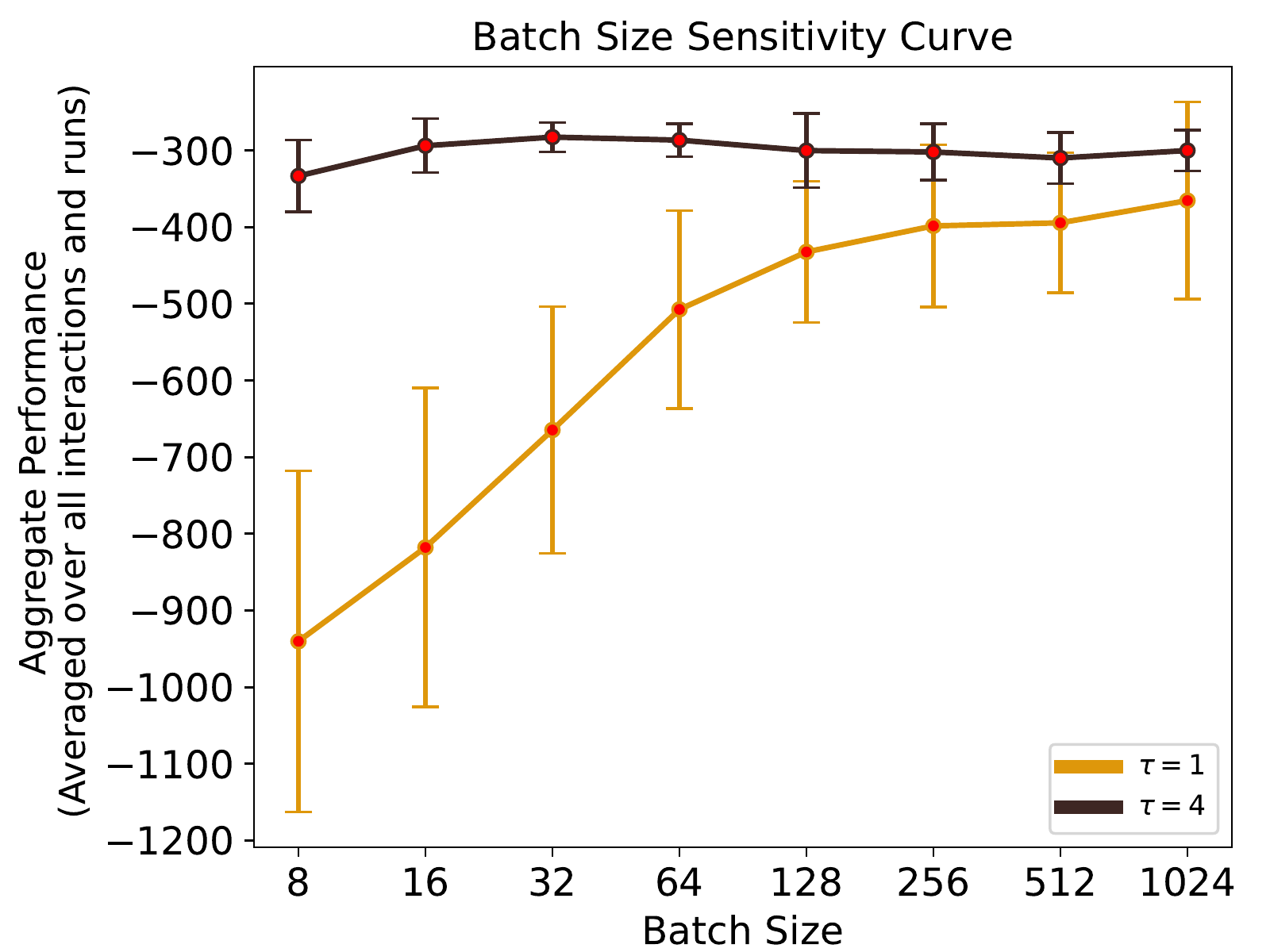}
        \subcaption{}
    \end{subfigure}%
    \begin{subfigure}{.23\textwidth}
        \centering
        \includegraphics[width=1\textwidth]{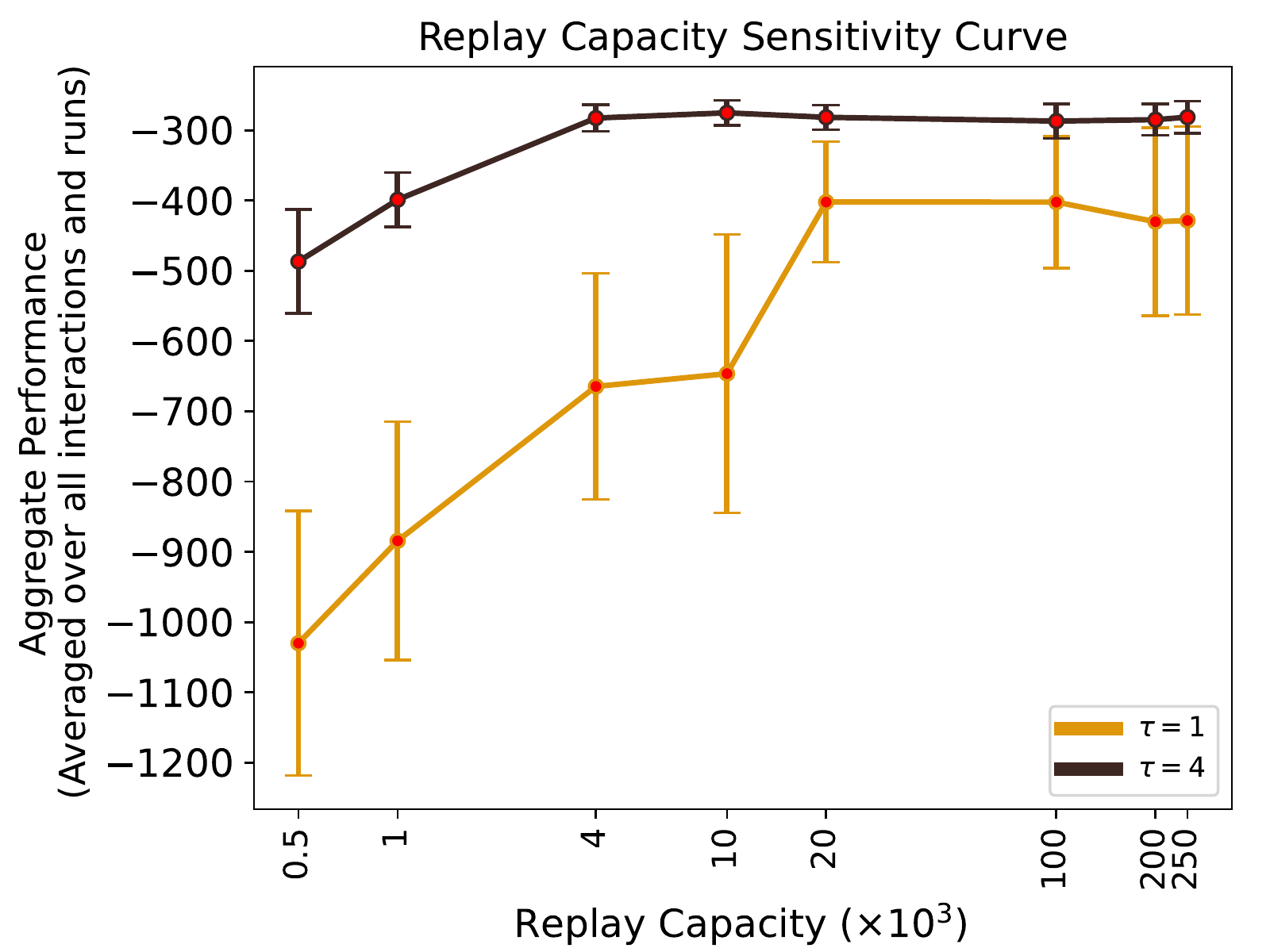}
        \subcaption{}
    \end{subfigure}%
    \begin{subfigure}{.23\textwidth}
        \centering
        \includegraphics[width=1\textwidth]{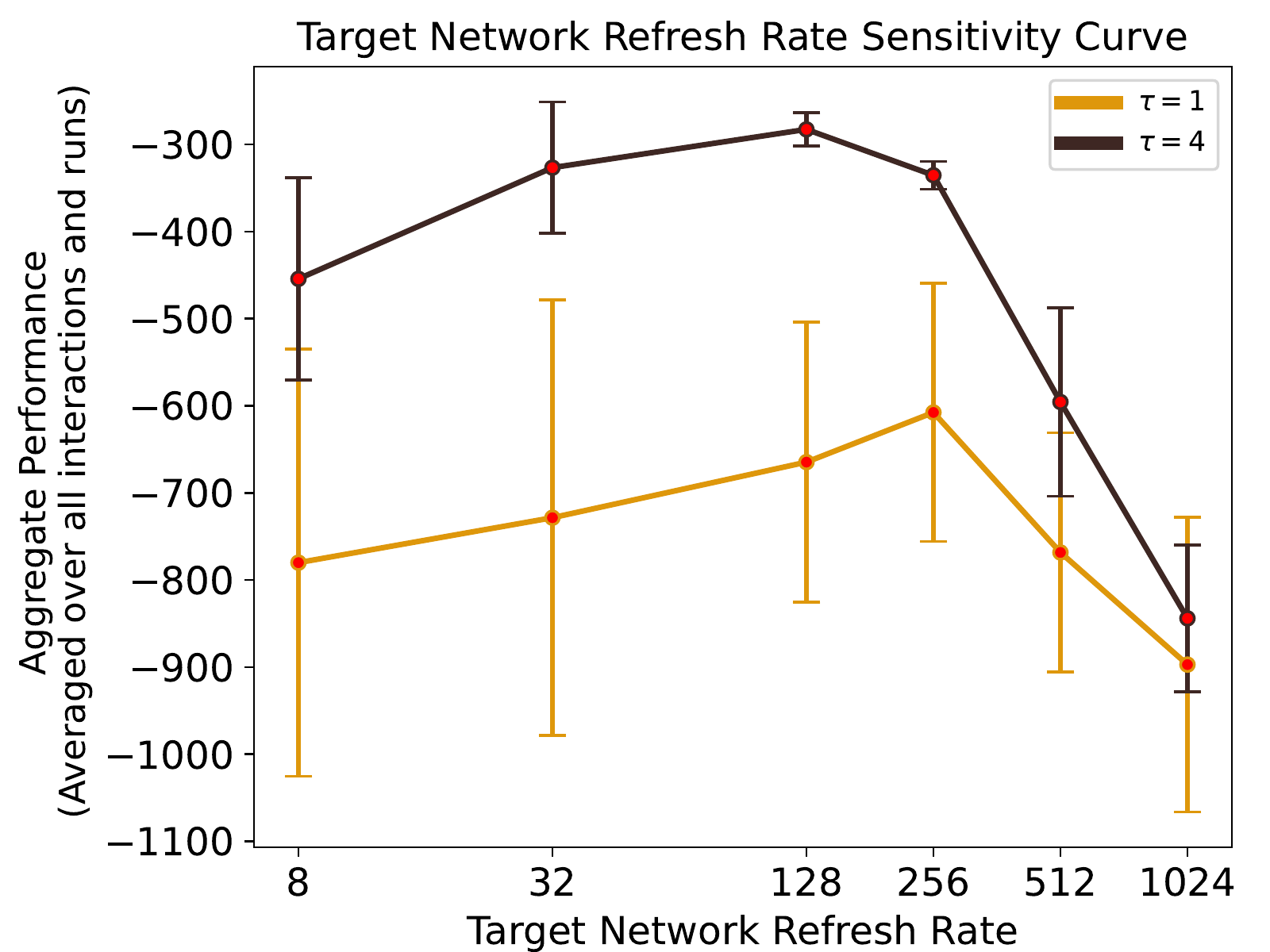}
        \subcaption{}
    \end{subfigure}
    
    \caption[short]{DQN hyperparameter sensitivity curves for $\tau=(1, 4)$ with $30$ runs: (a), (b), (c), and (d) show the sensitivity curve for learning rate, batch size, replay capacity, and target network refresh rate respectively. The error bars are computed using the Student-t distribution. The x-axis for all plots is on the log scale (base $2$).}
    \label{fig:dqn_sensitivity}
\end{figure*}

}
\begin{figure*}[!t]
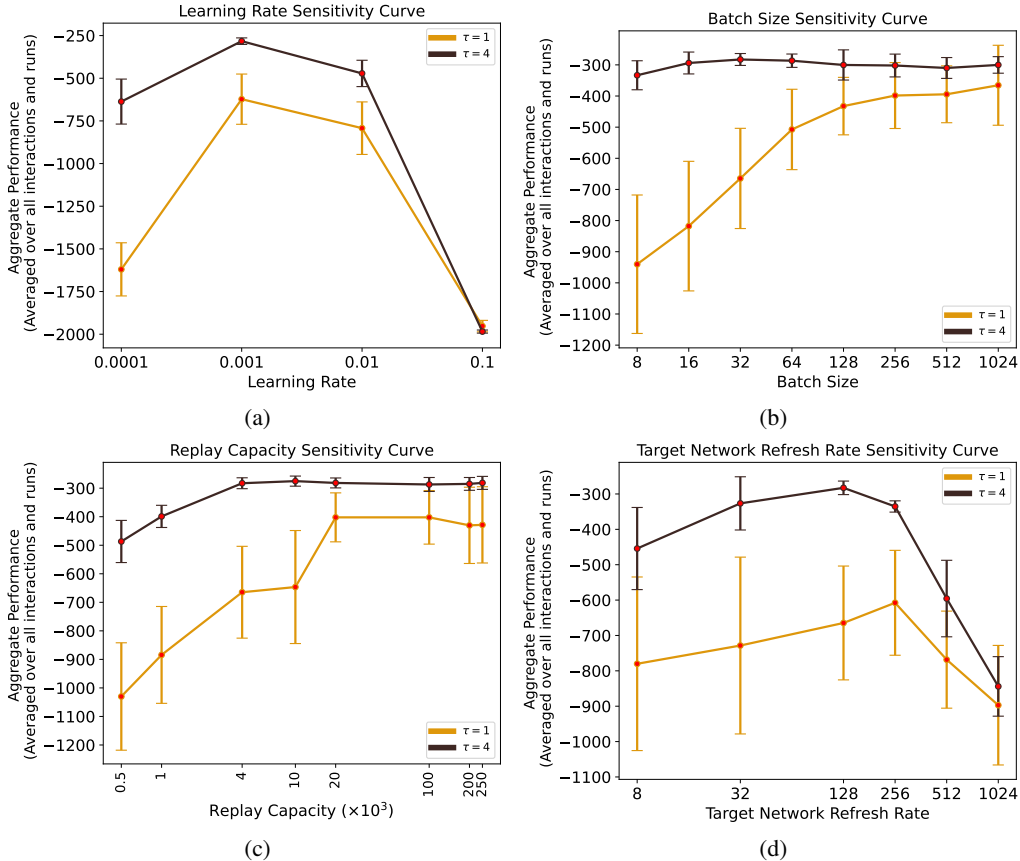

    \centering
    \begin{subfigure}{.40\textwidth}
        \centering
        \includegraphics[width=1\textwidth]{figs/lr_curve_runs30_student_t.pdf}
        \subcaption{}
    \end{subfigure}%
    \begin{subfigure}{.40\textwidth}
        \centering
        \includegraphics[width=1\textwidth]{figs/bs_curve_runs30_student_t.pdf}
        \subcaption{}
    \end{subfigure}

    \begin{subfigure}{.40\textwidth}
        \centering
        \includegraphics[width=1\textwidth]{figs/rc_curve_runs30_student_t.pdf}
        \subcaption{}
    \end{subfigure}%
    \begin{subfigure}{.40\textwidth}
        \centering
        \includegraphics[width=1\textwidth]{figs/tr_curve_runs30_student_t.pdf}
        \subcaption{}
    \end{subfigure}
    
    \caption[short]{DQN hyperparameter sensitivity curves for $\tau=(1, 4)$ with $30$ runs: (a), (b), (c), and (d) show the sensitivity curve for learning rate, batch size, replay capacity, and target network refresh rate respectively. The error bars are computed using the Student-t distribution. The x-axis for all plots is on the log scale (base $2$).}
    \label{fig:dqn_sensitivity}
\end{figure*}

\subsection{Hyperparameter Sensitivity Analysis}

In this section, we assess the variation in algorithm behavior when interpolating across different hyperparameters. For doing so, we use two-dimensional sensitivity curves where only one hyperparameter value is varied while keeping the others fixed~\cite{Patterson2020}. Sensitivity to a hyperparameter is assessed by how much the aggregate performance varies with a change in the hyperparameter value. Fig.~\ref{fig:dqn_sensitivity} shows the curves for four DQN hyperparameters, each for $\tau = (1, 4)$. We choose $\tau=4$ to compare against vanilla DQN ($\tau=1$). Our choice of $\tau=4$ is justified by fig.~\ref{fig:dqn_replay_curve}. $\tau=4$ provides a middle ground between good performance and the computation spent per step. Moreover, the uncertainty in the mean estimate is the lowest for $\tau=4$. 

One objective of the sensitivity experiments is to see if the better sample efficiency of higher $\tau$ values is only for the above specific hyperparameter setting or does it apply to a wide range of hyperparameter values. Knowing this is essential, especially for algorithm deployment in scenarios where hyperparameter tuning can be expensive. Another objective is to find the appropriate values of hyperparameters. As done for fig~\ref{fig:dqn_replay_curve}, we fix the total number of samples for each $\tau$ ($30$ runs each with $250,000$ interactions) and assess the aggregate performance with confidence estimates.

\subsubsection{Learning rate}

Fig.~\ref{fig:dqn_sensitivity} (a) shows the curve for the learning rate $\alpha$. We experiment with four values of $\alpha: (0.0001, 0.001, 0.01, 0.1)$. Our previous experiments used $\alpha=0.001$. For $\alpha=0.0001$, the learning is slow for both $\tau$ values. However, $\tau=4$ learns much faster than $\tau=1$. For $\alpha=(0.001, 0.01)$, the mean performance estimate for $\tau=4$ is higher than $\tau=1$. Moreover, the uncertainty estimates are low with $\tau=4$ with the lowest for $\alpha=0.001$. For $\alpha=0.1$, both $\tau$ values result in worse performance. This is because a large $\alpha$ takes overly aggressive gradient steps resulting in the agents learning nothing. Finally, if we look at the first three values of $\alpha$, $\tau=4$ is less sensitive (around the peak value at $\alpha=0.001$) to change in $\alpha$ than $\tau=1$, thereby making it relatively easier to choose an appropriate value of $\alpha$ for deployment. 

\subsubsection{Batch size}

Fig.~\ref{fig:dqn_sensitivity} (b) shows the curve for batch size $B$. We experiment with eight values of $B: (8, 16, 32, 64, 128, 256, 512, 1024)$. Our previous experiments used $B=32$. Note that $1024$ is the maximum value of $B$ that we can get without changing any other hyperparameter. This is because we fixed the replay start size to $1024$ and hence choosing $B$ greater than that would require changing the start size. However, we do not increase the replay start size to observe the sole effect of changing $B$ and to remain data efficient~\cite{VanHasselt2019}. The curve for $\tau=1$ increases continuously with the peak mean performance at $B=1024$. However, the uncertainty estimates of $\tau=1$ are very high for all values of $B$, thereby making it difficult to choose the appropriate value of $B$. On the other hand, $\tau=4$ is relatively less sensitive to a change in $B$ and has better mean performance with low uncertainty, which lets us choose $B=32$ that has the best mean performance and lowest uncertainty. 

One can argue that choosing a large value of $B$ with $\tau=1$ may provide benefits similar to increasing $\tau$. However, this is not true because for a large $B$, the worst value (lower end of the confidence interval) of the mean estimate for $\tau=1$ is considerably lower than the worst value for $\tau=4$. Moreover, choosing a large $B$ may not always be feasible due to memory constraints as it requires loading more samples at each step~\cite{memory_constraint}. Increasing $B$ also requires more computation per step. We argue that it is instead wiser to spend more computation on increasing $\tau$ while keeping $B$ small. This is clear from the curve for $\tau=4$ where $B=(32, 64)$ are the best performing values. 

Along similar lines, it is interesting to compare $\tau=1, B=32$ with $\tau=4, B=8$. Both use $32$ samples to make parameter updates with the difference in the way they update. The first one uses all $32$ randomly sampled transitions at once, while the second randomly samples $8$ transitions and uses them to make an update, doing this sequentially four times at each step. It is clearly visible that the worst mean performance for the second is much better than the best mean performance for the first. This traces back to connections of replay with planning where putting a loop around the model-based update results in better performance without consuming additional data~\cite{VanHasselt2019}. 

\subsubsection{Replay capacity}

Fig.~\ref{fig:dqn_sensitivity} (c) shows the curve for the replay capacity $\Omega$. We experiment with eight different values of $\Omega: (0.5, 1, 4, 10, 20, 100, 200, 250) \times 10^{3}$. Our previous experiments used $\Omega=4000$. When $\Omega$ is too small, the life of a transition in the buffer reduces as it gets discarded to make room for new transitions. Hence the replay buffer contains more transitions from a recent policy. This can have a negative effect as there is less chance of breaking correlations between the transitions used for updates. When $\Omega$ is too large, the updates can use transitions from an old policy which can be distributed in parts of the state space irrelevant to solving the task. In the extreme case, $\Omega$ can be logically equal to the total number of interactions with the environment, in which case it will not forget any experience ($250,000$ in our experiments). 

The curve shows that the performance is worst when $\Omega$ is relatively small. For $\tau=4$, $\Omega=(500, 1000)$ result in inferior performance with high uncertainty in the mean estimates. However, for $\Omega=4000$ and greater, the performance is less sensitive to a change in $\Omega$ with relatively low uncertainty. $\tau=4$ with $\Omega=10,000$ gives the highest mean performance with the lowest uncertainty in the estimate. Note that the mean performance and uncertainty estimates of $\tau=4$ are better than $\tau=1$ for all $\Omega$ values, with the difference clearly visible for $\Omega$ less than $20,000$. For $\tau=1$, the mean performance increases till $\Omega=20,000$, after which it degrades slightly. However, high uncertainty in the mean estimates makes it difficult to pick an appropriate value of $\Omega$ for $\tau=1$. %The better performance, relatively low uncertainty estimates, and less sensitivity to a change in $\Omega$ with $\tau=4$ make it easier to pick an appropriate value of the hyperparameter for $\tau=4$ than $\tau=1$. 

It is interesting to note that the nature of the replay capacity sensitivity curve is quite different for $\tau=1$ and $\tau=4$. The same holds for the batch size sensitivity curve from the previous subsection. $\Omega, B,$ and $\tau$ all are hyperparameters of experience replay and they may be interacting with each other in a non-trivial manner. Our two-dimensional sensitivity curves indicate a change in performance with variation in a single hyperparameter but do not capture interactions between multiple hyperparameters. It will be interesting to work on interacting hyperparameters in the future to get a deeper insight into the effects of increasing $\tau$. 

\subsubsection{Target network refresh rate}

Fig.~\ref{fig:dqn_sensitivity} (d) shows the curve for the target network refresh rate $C$. We experiment with six values of $C: (8, 32, 128, 256, 512, 1024)$. Our previous experiments used $C=128$. The frequency with which the lagging target network is refreshed affects the stability and performance of DQN. When $C$ is too small, the delay between the time when the target is computed and the time when parameters are updated decreases, thereby causing oscillations or divergence of policy. However, when $C$ is too large, the target is computed using a very old policy that may be very different from the current policy, which may cause inconsistency in the parameter update. Our results agree with this: both $\tau=(1, 4)$ seem sensitive to a change in $C$ with a worse performance when $C$ is too small or too large. 

It is interesting that a small $C$ has a \textit{milder} effect on performance than a large $C$ for $\tau=4$. On the other hand, both extremes of $C$ result in similar performance drops for $\tau=1$. This hints that $\tau=4$ is able to handle non-stationary targets better than $\tau=1$. This might be because $\tau=4$ reuses the data more to improve the Q-network's approximation of the value function faster than $\tau=1$. Subsequently, refreshing the target network more frequently (small $C$) causes the target to become more accurate faster. When updating the parameters more frequently ($\tau=4$) with this better target, the drop in performance due to target oscillations is lesser.

The highest mean performance occurs at $C=128$ for $\tau=4$ and at $C=256$ for $\tau=1$. However, the best value of $C$ is not clear for $\tau=1$ because of the high uncertainty in the mean estimate. The uncertainty is high for all values of $C$ for $\tau=1$ with the highest when $C$ is small. For $\tau=4$, it is easy to pick the appropriate value of $C$: $128$ is the best-performing value with relatively low uncertainty. %However, it is much easier to choose an appropriate $C$ for $\tau=4$ because of relatively confident high performance. 

\section{Conclusions}

In this work, we investigated how varying the replay frequency affects DQN's performance, sample efficiency, and sensitivity to hyperparameters in the Mountain Car environment. To validate our hypothesis, we experimented with different replay frequencies, measured the variability in performance, and tested with different hyperparameter values. The empirical results suggest that (1) increasing $\tau$ results in better sample efficiency than vanilla DQN ($\tau=1$); (2) DQN with higher $\tau$ values generally gives better mean performance with tighter confidence and tolerance intervals; (c) higher $\tau$ makes DQN less sensitive to other hyperparameters, thereby easing the task of hyperparameter selection.

%Moreover, we found similar results with double DQN agents.

\comment{
In this work, we investigated how the choice of replay frequency affects a deep RL agent’s performance and sample efficiency. To test our hypothesis---increasing replay frequency makes learning faster---we considered using two deep RL algorithms (DQN and DDQN) in the Mountain car environment. Our preliminary results suggest that increasing the replay frequency can make learning faster early on during training for the DQN agents, i.e. they achieve the same level of performance in a lesser number of environment interactions. In the future, we plan to:

\begin{itemize}
    \item Improve the computational efficiency. A major drawback of increasing replay frequency is that it requires more computation per step. In our experiments, the replay frequency was constant throughout training. However, the learning curves suggest that higher replay frequency has a dominant effect only early on in the training. Hence once a near-optimal performance is reached, it might be possible to switch to the traditional replay frequency of $1$. However, we would need to check if there is a performance drop due to this. Moreover, it can be challenging to identify the exact optimal point to make the switch during training.
    \item Assess whether increasing replay frequency can lead to a further improvement in performance, given a limited number of samples (environment interactions). 
    \item Include more number of runs; empirically test the hypothesis for DDQN and compare it with DQN.
    \item Assess the interaction between multiple hyperparameters. In this draft, we froze all other hyperparameters to observe the difference by changing just the replay frequency. However, the optimal setting of other hyperparameters with replay frequency $\tau_1$ may not be the same as with $\tau_2$. Hence we need to consider how multiple hyperparameters interact with each other. In particular, we plan to consider the hyperparameters that can have a significant effect on the algorithm like the learning rate, target refresh rate, and batch size. 
    \item Use other potentially informative methods to convey results than online returns like worst-case performance, behavioral metrics like state visitation, and area under the curve. It might also be interesting to evaluate the effect of increasing replay frequency offline.
    \item Run experiments without Xavier initialization to observe if its presence in the above experiments causes an additional positive effect when increasing replay frequency. 
\end{itemize}}
%• specify the hypotheses and questions you seek to answer
%From the cookbook: If the goal is to understand the utility of a modification to the algorithm or to compare two algorithmic ideas or to just compare two algorithms, then we SHOULD NOT use default HPs.

%\section{Future Plan}
%assess the cross product of n settings of the replay
%capacity (from __ to __ ) and m settings of the batch size (from __ to __), 

%\section{Acknowledgement}

%\section{Electronic Submission}
%\label{submission}

%\begin{itemize}
%\item Submissions must be in PDF\@.
%\item Submitted papers can be up to eight pages long, not  
%\item \textbf{Do not include author information or acknowledgements} in your
%\end{itemize}

%\medskip

%\textbf{Graphics files} should be a reasonable size, and included from an appropriate format. 
%Use vector formats (.eps/.pdf) for plots,
%lossless bitmap formats (.png) for raster graphics with sharp lines, and jpeg for photo-like images.

%The style file uses the \texttt{hyperref} package to make clickable links in documents. If this causes problems for you, add
%\texttt{nohyperref} as one of the options to the \texttt{icml2021} usepackage statement.

%\subsubsection{Paragraphs and Footnotes}
%You can use footnotes\footnote{Footnotes
%should be complete sentences.} 
%\footnote{Multiple footnotes can
%appear in each column, in the same order as they appear in the text, but spread them across columns and pages if possible.}

%\subsection{Figures}

%(use the environment \texttt{figure*} in \LaTeX). Always place
%two-column figures at the top or bottom of the page.

\section*{Software}
The code of our experiments is available at \href{https://github.com/animeshkumarpaul/IncreasingReplay}{https://github.com/animeshkumarpaul/IncreasingReplay}.

% Acknowledgements should only appear in the accepted version.
\section*{Acknowledgements} 

We took the initial codebase from \href{https://github.com/dongminlee94/deep_rl}{Dongmin repository}. Thanks to \href{https://andnp.github.io/}{Andy} for sharing the code for tolerance intervals.

% In the unusual situation where you want a paper to appear in the
% references without citing it in the main text, use \nocite
%\nocite{langley00}

\bibliography{main}
\bibliographystyle{icml2021}

\end{document}